\documentclass[runningheads]{llncs}

\usepackage{eccvabbrv}

\usepackage{graphicx}
\usepackage{booktabs}

\usepackage[accsupp]{axessibility}  

\usepackage{orcidlink}
\usepackage{amsmath}
\usepackage{multirow}
\usepackage{subcaption}
\usepackage{array}
\usepackage{float}
\usepackage[symbol]{footmisc}

\newcommand\rgt{\aftergroup\mathclose\aftergroup{\aftergroup}\right}

\begin{document}

\title{Latent-INR: A Flexible Framework for Implicit Representations of Videos with Discriminative Semantics} 

\titlerunning{Latent-INR}

\newcommand*\samethanks[1][\value{footnote}]{\footnotemark[#1]}

\author{Shishira R Maiya$^{*}$  \and
Anubhav Gupta$^{*}$  \and
Matthew Gwilliam \and
Max Ehrlich \and
Abhinav Shrivastava
}

\authorrunning{Maiya et al.}

\institute{University of Maryland, College Park, USA }

\maketitle
\def\thefootnote{*}\footnotetext{Equal contribution}

\begin{abstract}
    Implicit Neural Networks (INRs) have emerged as powerful representations to encode all forms of data, including images, videos, audios, and scenes. 
    With video, many INRs for video have been proposed for the compression task, and recent methods feature significant improvements with respect to encoding time, storage, and reconstruction quality.
    However, these encoded representations lack semantic meaning, so they cannot be used for any downstream tasks that require such properties, such as retrieval.
    This can act as a barrier for adoption of video INRs over traditional codecs as they do not offer any significant edge apart from compression. 
    To alleviate this, we propose a flexible framework that decouples the spatial and temporal aspects of the video INR.
    We accomplish this with a dictionary of per-frame latents that are learned jointly with a set of video specific hypernetworks, such that given a latent, these hypernetworks can predict the INR weights to reconstruct the given frame. 
    This framework not only retains the compression efficiency, but the learned latents can be aligned with features from large vision models, which grants them discriminative properties.
    We align these latents with CLIP and show good performance for both compression and video retrieval tasks.
    By aligning with VideoLlama, we are able to perform open-ended chat with our learned latents as the visual inputs.
    Additionally, the learned latents serve as a proxy for the underlying weights, allowing us perform tasks like video interpolation. 
    These semantic properties and applications, existing simultaneously with ability to perform compression, interpolation, and superresolution properties, are a first in this field of work.
\end{abstract}

\section{Introduction}

In today's age of content explosion, large quantities of data are created every second, and storing them reliably and efficiently is of utmost importance for many applications. 
A scalable compression technique enables companies to provide better services at reduced cost and helps the end consumer by improving their access to high-fidelity data in addition to decongesting the network. 
Since the early 90s, several compression techniques have been created and widely deployed for this exact purpose. 
Out of these, JPEG \cite{wallace1991jpeg} for images, HEVC \cite{sullivan2012overview}, AV1\cite{chen2018overview}, and H.264~\cite{H264} for videos have emerged as the most popular choices, owing to their simple design and scalable performance. 

In the past decade, the rise of deep learning led to a renaissance in computer vision, eventually impacting the visual data compression landscape \cite{Ehrlich_2019_ICCV,Ball2018VariationalIC,Mentzer2020HighFidelityGI}. 
Despite their success, these ML-based codecs have not seen widespread adoption like traditional codecs. 
This is in part due to failure to generalize, since ML codecs trained on large datasets can give sub-optimal compression for data points that differ significantly from their training set \cite{Zhang2021OnTO,DBLP:conf/aaai/CaoJLWY22}. 
Implicit Neural Representations (INR) attempt to avoid the generalization issue by operating internally. 
Instead of training large models that learn to identify \textit{general} patterns in training data and apply them to specific out-of-distribution data, implicit techniques involve training a small model to exploit the \textit{specific} patterns for the given data point. 
That is, for video compression, this approach would train one network per video, and for image compression, it would train one network per image.
The resulting model is essentially a function that represents the underlying signal in spatial/temporal space. 

\begin{figure}[t]
    \centering
    \includegraphics[width=\linewidth]{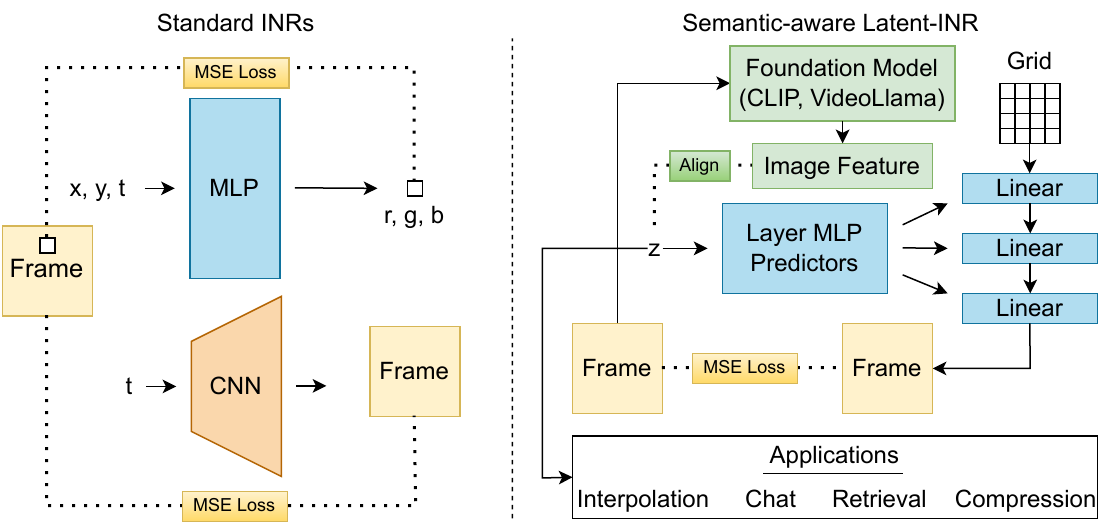}
    \caption{\small{Existing INRs for video (left) typically take some time-coordinate, or time and positional coordinates and train a single network to reconstruct a video. In contrast to these, we propose an INR system where a dictionary of implicit latent codes is learned for a video, one latent per frame. The latents are aligned to the image features of a large vision model, while simultaneously an INR system is learned which, given these latent codes, generates a positional INR which can reconstruct the frame. With this framework, we successfully develop an INR which performs both reconstructive tasks like compression, and semantic downstream tasks like retrieval and interactive chat.}}
    \label{fig:teaser}
\end{figure}


Despite these advances, neural video compression remains unsolved. 
Various methods address issues of compression quality~\cite{chen2021nerv,kim2022scalable}, but 
two crucial questions remain unanswered -- (i) how to scale for longer videos given architectural rigidity and (ii) how to reduce long encoding time due to training a network for every video.
Although recent works make some progress for these~\cite{maiya2023nirvana},
the training time is still quite long, and INR behavior for lossy compression is not well-understood~\cite{Padmanabhan_2024_CVPR}, limiting potential for practical adoption.

Furthermore, these approaches for INR tackle only one axis of the problem, i.e., how to formulate video INRs with the primary goal of compression. 
These aim to solve problems like long encoding time directly, by reducing it.
In contrast to these works, we instead aim to justify the compute and time needed to train implicit representations.
So, as a step towards ML-based codecs with compelling real-world potential, we present Latent-INR -- a new flexible framework for formulating video INRs, where in addition to compression, the INR enables downstream tasks like retrieval and video question answering, without the need to decode the video.
Our framework consists of two parts: (i) a dictionary of learnable latents, one for each frame, and (ii) a set of hypernetworks learned on the entire video which, given a latent as input, predict frame-specific weight modulations on the shared base network.
This shared base takes a spatial coordinate grid as input and outputs the specific frame

This design allows us to separate the spatial and temporal aspects of the video by modeling them separately. 
We can view the set of hypernetworks as a base model that learns the general structure and style of the video, while each learned latent conditions it to output a specific frame. 
The latent here acts as a proxy for the weights of the frame-specific INR.
This property is apparent from the video interpolation ability of our model - a task that other video INR representations struggle to perform. 
Like other video INRs, our method is competitive for compression, but uniquely retains the properties of original coordinate-based INR.
That is, our continuous representations of frames allows for spatial interpolation, which can be leveraged for superresolution and a decoding paradigm we refer to as ``any-resolution inference.''
That is, at inference/decoding time, our same model, with no changes to latents or architecture, can decode a video at any resolution - a key feature missing from traditional codecs.
This latent is also quite flexible, and according to the procedure shown in Figure~\ref{fig:teaser}, we can align it with the features from a large vision model, such as CLIP~\cite{Radford2021LearningTV} to encode the visual semantics of the frame while retaining nice properties such as alignment with CLIP text embeddings.
This allows for a whole spectrum of applications, including frame, concept, and whole video retrieval with text queries.

In summary, our framework gives that extra edge apart from compression to ML-based codecs, paving the way for their widespread adoption. Concretely, 
\begin{itemize}
    \item We propose an auto-decoder latent-based framework with spatio-temporal decoupling for implicit video representations. Compared to other video INR methods, this is a new way of formulating the problem. 
    \item Our system has good compression performance, competing well with other ML-based codecs for PSNR, BPP, and decoding speed while also enabling any-resolution inference.
    \item  The learnt latent embeddings from our framework demonstrate internal generalization from the encoded dataset, achieving video interpolation, a task that other INR based methods struggle to achieve. 
    \item  We align our latents with large foundational models like CLIP \cite{Radford2021LearningTV}, thus making our representations useful for retrieval tasks. 
    \item We align our entire dictionary with video features for VideoLlama~\cite{damonlpsg2023videollama} to enable chat-style applications, including video question answering and captioning.
\end{itemize}

\section{Related Work}
\label{sec:related_work}

\textbf{Implicit Neural Representations (INR's)} are a class of neural networks designed with the intention of representing a given data point or dataset perfectly rather than exploiting general patterns and generalizing for unseen data. 
SIREN \cite{sitzmann2020implicit} pioneered the use of periodic activations to train simple MLP's that worked well across images, SDF and audio. 
This was followed by a host of works that improved the training process of INR's by making them faster \cite{Saragadam2023WIREWI,tancik2020fourfeat,mueller2022instant} work across multiple scales \cite{DBLP:journals/corr/abs-2202-03532} and encode multiple data points \cite{Dupont2022FromDT}. 
Models that used meta learning \cite{Strmpler2021ImplicitNR,tancik2020meta} started gaining ground as they offered the advantages of compression along with generalization. 
\cite{schwarz2022metalearning,tack2023learning} further made improvements to this line of work by directly learning sparse-INR's leading to improved compression and improved optimization by dataset selection respectively. 

\textbf{Hypernetworks} are a class of networks optimized for predicting parameters of another network, with the aim of generalizing across unseen tasks\cite{Finn2017ModelAgnosticMF}. 
Some utilized these for scenes~\cite{sitzmann2019scene,sitzmann2021light,chiang2022stylizing}.
Trans-INR \cite{chen2022transinr} introduced the paradigm of using a transformer based hypernetwork to convert data directly from image-space to INR's. \cite{kim2023generalizable} improved upon this idea and made the important observation that it is sufficient to modulate only the first hidden layer of an INR to represent a dataset of points. 
Unfortunately, these hypernetworks act on input data points which require test-time optimizations, making them unsuitable for compression tasks. 
\cite{sen2022inr} try to overcome this with an ``auto-decoder'' framework, where learnable latents represent a dataset of videos, with each latent corresponding to a single video, such that no encoder is needed. 
Others have investigated this paradigm for a variety of modalities\cite{sen2023hyp,schwarz2023modality,bauer2023spatial}.
Still, the lack of decoupling space from time prohibits the method from scaling to real-world videos. 

\textbf{Video INRs} have recently gained popularity for compression. 
\cite{chen2021nerv} was the first implicit representation which modelled a video as a function mapping the temporal coordinates to the corresponding frames. 
Later works \cite{li2022nerv,bai2023ps,chen2023hnerv,he2023towards} iterated on this method, providing improvements in performance. 
\cite{kim2022scalable} enhanced this concept by incorporating hash-grid~\cite{mueller2022instant} representations to speed up encoding times. 
NIRVANA \cite{maiya2023nirvana} represented a video using a series of smaller INR models trained in an autoregressive manner to scale for longer videos. 

\textbf{Video Interpolation} has been a fundamental task in computer vision, helping in creating smoother visual experiences.
Over the past few years, deep learning based methods have vastly improved the quality of these interpolations \cite{Shi2016RealTimeSI,Jiang2017SuperSH}.However, current INR-based video encoders lack this feature (see discussion in \cite{hao2022cnerv,chen2023hnerv}, for example), hindering their widespread usage.

\textbf{Video Retrieval } is an essential process in the digital media landscape, where the objective is to efficiently search and extract specific video content from expansive datasets. The complexity of understanding and indexing diverse video content has traditionally posed significant challenges. However, with the advent of machine learning-based methods, there has been a remarkable improvement in both the accuracy and efficiency of video retrieval systems \cite{2019ASF,Bain2021FrozenIT,Luo2021CLIP4ClipAE}.
These advances are limited to systems requiring an additional model, which can act as a burden on the system as they do not compress the data. 


\begin{figure}[t]
    \centering
    \includegraphics[width=\textwidth]{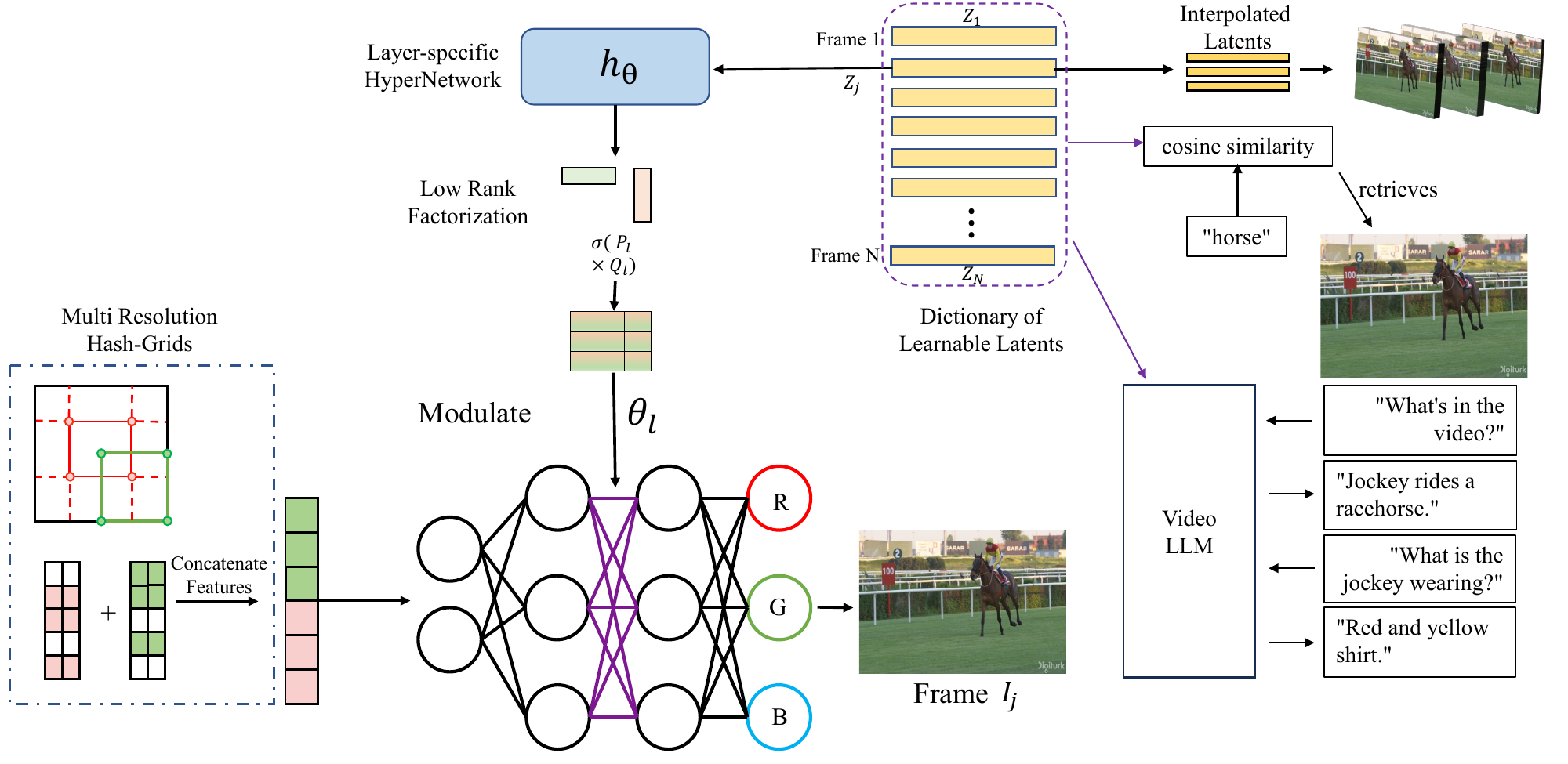}
    \caption{\small{We propose a new framework for video INR models by decoupling the spatial and temporal aspects of modeling. Our framework consists of auto-decoder based learnable latents that modulate the base network using a hypernetwork, via low-rank modulation. Once encoded, the resulting latents act as a proxy for the underlying weights of the representation. On the right, we show the use of these latents for additional tasks like video interpolation. By aligning these latents to the embedding space of foundational models like CLIP and VideoLlama, we also perform retrieval and chat. 
    }}
    \label{fig:architecture}
\end{figure}

\section{Approach}
\label{sec:approach}
\subsection{Background}

Implicit Neural Representations parameterize a function,
\begin{equation*}
    f_{\theta}:X \rightarrow Y \quad \textrm{where} \quad X = \left\{ (x_i, y_i) | 0 \leq x_i \leq W,  0 \leq y_i \leq H \right\}
\end{equation*}
which represents a mapping between the coordinate space, with \textit{height} H and \textit{width} W, and the underlying signal $Y$.
This formulation is usually trained with a standard MSE-loss: $||f_{\theta}(X) - Y||_{2}$. For a given video $V \in R^{N \times H \times W \times 3}$ containing $N$ frames, \cite{sitzmann2020implicit} represents them as pixels moving across time, i.e.,
\begin{equation*}
    f_{\theta}(x,y,t) = Y_t
\end{equation*}


Other formulations exist which learn frame-based~\cite{chen2021nerv} or patch-based~\cite{maiya2023nirvana} representation, yet in each of these formulations, the focus is on representing the underlying data, with the added motivation of compressing it. However, none of these systems are designed with the goal of making these representations, $f_\theta$, useful for downstream tasks~\cite{Padmanabhan_2024_CVPR,Papa_2024_CVPR}. Instead, we utilize a learnable latent, $z$, as a part of an auto-decoder framework, along with a hypernet $h$ to not only compress but to create useful representations. 
\begin{equation}
\begin{aligned}
    f_{\theta}((x,y)|\theta_{t}) = Y_{t} \quad
    \theta_{t} = h(z_{t})
\end{aligned}
\end{equation}
The resulting latent $z$ can be used for various downstream tasks like interpolation and retrieval, as we show in our work.

\subsection{Latent-INR}
Directly predicting the weights $\theta$ of the base network $f$, using the hypernet $h$, is expensive, parameter-heavy, and unsuitable for compression. Hence, we follow \cite{skorokhodov2021adversarial} \cite{schwarz2023modality} and instead predict low-rank matrices, which are then applied to the base network weights. This type of modulation acts as a form of subnetwork selection, analogous to systems proposed in \cite{frankle2018lottery} \cite{ramanujan2020s}. For a base network $f$ with $L$ layers, our formulation now looks like 
\begin{equation}
\begin{aligned}
    f_{\theta}((x,y)|\theta^{l_{1}}_{t},\theta^{l_{2}}_{t} ... \theta^{l_{L}}_{t}) = Y_{t} \\
    \theta^{l}_{t} = \sigma ( P^{l} \times Q^{l}) \cdot \theta^{l} \quad
    h_{l}(z_{t}) = [P^{l},Q^{l}]
\end{aligned}
\end{equation}
where $\theta^{l}$ represents the weights of the $l$-th layer and $\theta^{l}_{t}$ denotes the modulated weights for frame $t$.  
Here, $\sigma$ signifies an activation function on the matrix-product of low rank matrices $P^{l}$ , $Q^{l}$, which are of dimensions $R^{N \times r}$ and $R^{M \times r}$, where $N \times M$ is the width of the base network $f_{\theta}$ and rank $r \ll (N,M)$. 
These matrices are responsible for adjusting the weights $\theta_{l}$ as dictated by the corresponding hypernetwork $h_{l}$. Note that all hypernetworks use the same latent $z_{t} \in R^{D}$  as input. The rank $r$ and the number of modulated layers essentially act a hyperparameters that control the compression-performance trade-off. \\
\subsection{Model architecture}
In our experiments, both the base network $f_{\theta}$ and hypernetworks $h_{l}$  are feedforward MLP's that take in a coordinate input. Following \cite{maiya2023nirvana}, we also propose a variation to the base network with an additional convolutional up-sample block, which accepts coordinates of centroids as input and gives frame patches as output. We use the standard ReLU for base network and tanh for the hypernetwork as the respective non-linearities. 
The latents $Z$ are initialized to be a standard normal with small variance, as we found empirically that this made the convergence faster. The complete model architecture is presented in Figure \ref{fig:architecture}. For more details, see Appendix.

\subsection{Model Compression}
We train this entire system end-to-end with MSE-loss as the objective function. Once trained, we apply a standard quantization to all network parameters, further reducing the required storage. Given $\phi$, a flattened parameter tensor, we transform it according to the following equations
\begin{equation}
    \phi_{i} =  \left\lceil \frac{\phi_{i} - \phi_{\text{min}}}{2^{b}} \right\rfloor \quad 
    \text{scale} = \frac{\phi_{\text{max}} - \phi_{\text{min}}}{2^{b}}
\end{equation}
where the $\lceil\cdot\rfloor$ (round) operation converts its argument to the nearest integer as dictated by bit width $b$ of the quantization process. We also store the scale, $\phi_{\text{max}}$, $\phi_{\text{min}}$ and the parameter shapes. These quantized values for all parameters are concatenated and further compressed using Huffman encoding.

\subsection{Interpolation}
Given a video of $N$ frames and a scale $\alpha$, the task of interpolation involves creating $\alpha \cdot N$ coherent frames. Once we encode a video using our framework, we perform linear interpolation on the frame latents $\{z_{t}\}$ 
and pass the resulting latent through the hypernetwork. This gives us the weight modulation required in the INR, and the updated base network is used to obtain the interpolated frames: 
\begin{equation}
\begin{aligned}
    z_\text{inter} &= \beta_i \cdot z_{t} + (1 - \beta_i) \cdot z_{t-1} \quad
    Y_\text{inter} &= f_{\theta}(X; h(z_\text{inter}))
\end{aligned}
\end{equation}

where, 
\begin{equation*}
    \beta_i \in \left[\frac{1}{\alpha}, \frac{2}{\alpha}, . . ., \frac{\alpha - 1}{\alpha}\right]
\end{equation*}
essentially generating $\alpha - 1$ frames between any two given frames. We train with held out frames and show results for $\alpha \in \{2,4,8\}$.

\subsection{Downstream Tasks}
\label{subsec:retrieval}

\noindent\textbf{Retrieval.} Video retrieval involves searching and retrieving videos or clips from a large database based on similarity to given user search queries that are usually in the form of text. 
This can be viewed as a function $R$ mapping query $q$ to a set of corresponding videos $V$.
\begin{equation}
    \begin{aligned}
        R: q \rightarrow V
    \end{aligned}
\end{equation}

The function $R$ can use any similarity measure like cosine, euclidean, or nearest neighbors to retrieve matches. We encode a dataset of videos using our Latent-INR framework and use the resulting trained latents as our frame level representations. To ensure these latents share the same space as the text queries, we add a cosine similarity loss between the latents and the CLIP image embeddings of the corresponding frames. Our encoding loss function is modified to be:
\begin{equation}
    \begin{aligned}
        L = L_{\text{MSE}} + \lambda \cdot L_{\text{clip}}(Z_{t},Z^{\text{clip}}_{t})
    \end{aligned}
\end{equation}
where $Z^{\text{clip}}_{t}$ is the CLIP Image embedding of the input frame and $\lambda$ controls the strength of this loss. In all our experiments, $\lambda$ is set to 0.01.

\noindent\textbf{Chat.} We modify the formulation from retrieval slightly, aligning our dictionary of features to VideoLlama~\cite{damonlpsg2023videollama} instead of CLIP.
Since the shapes are not compatible, we treat our latents as tokens and project the dimension to match the VideoLlama space.
With this, we are able to integrate our latents with a powerful LLM, substituting our latents for the raw video input tokens.
We can then perform any task that VideoLlama can, in particular question answering and captioning.
We wish to emphasize that our latents are flexible -- we can align well with any large off the shelf model, for any downstream task.

\begin{figure}[t]
    \centering
    \includegraphics[width=\textwidth]{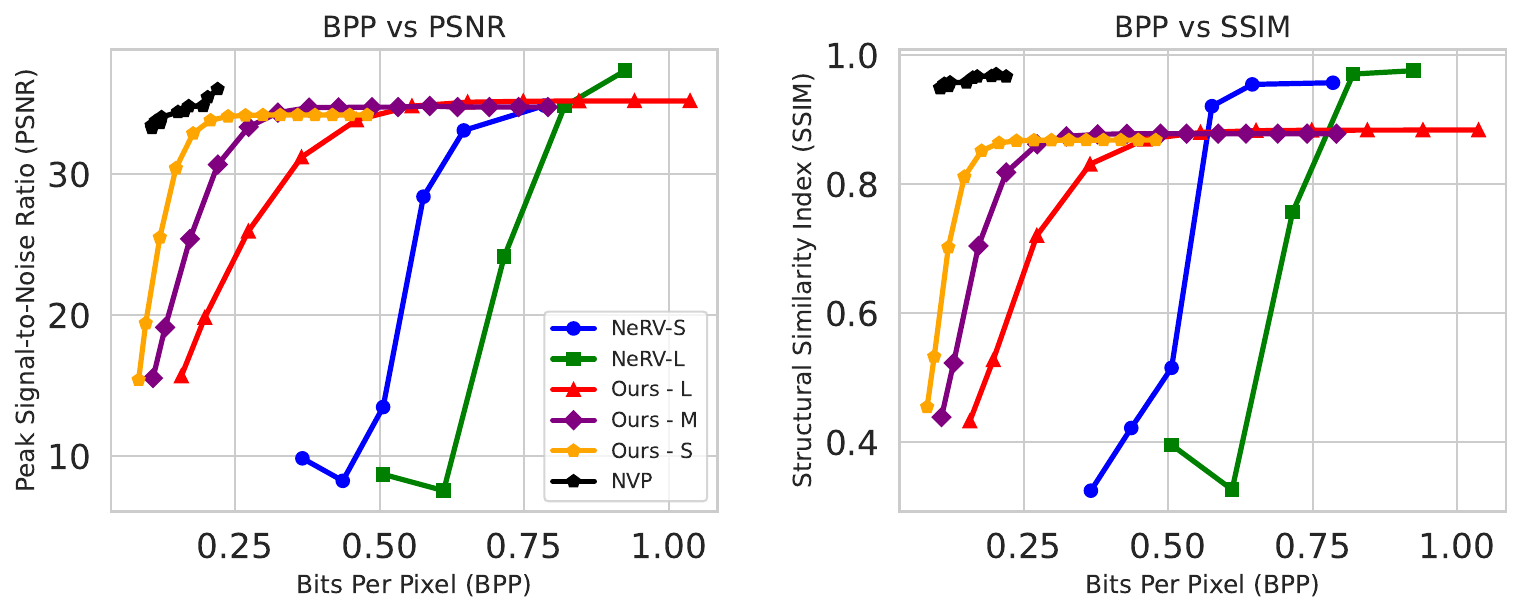}
    \caption{\small{We plot the rate distortion curves on PSNR and SSIM to compare compression with other methods. We observe that our large model achieves comparable PSNR to the current SOTA~\cite{kim2022scalable}}. Note that, while not plotted here, our decoding FPS is superior. Additional per-video results are available in the Supplementary.}
    \label{fig:rate_distortion}
\end{figure}

\section{Experiments}

\subsection{Video Compression}
We perform comparative analysis for video compression on the standard Ultra Video Group (UVG) dataset~\cite{10.1145/3339825.3394937}. 
\begin{figure}[ht]
  \begin{minipage}{0.49\linewidth}
    \centering
    \includegraphics[width=\linewidth]{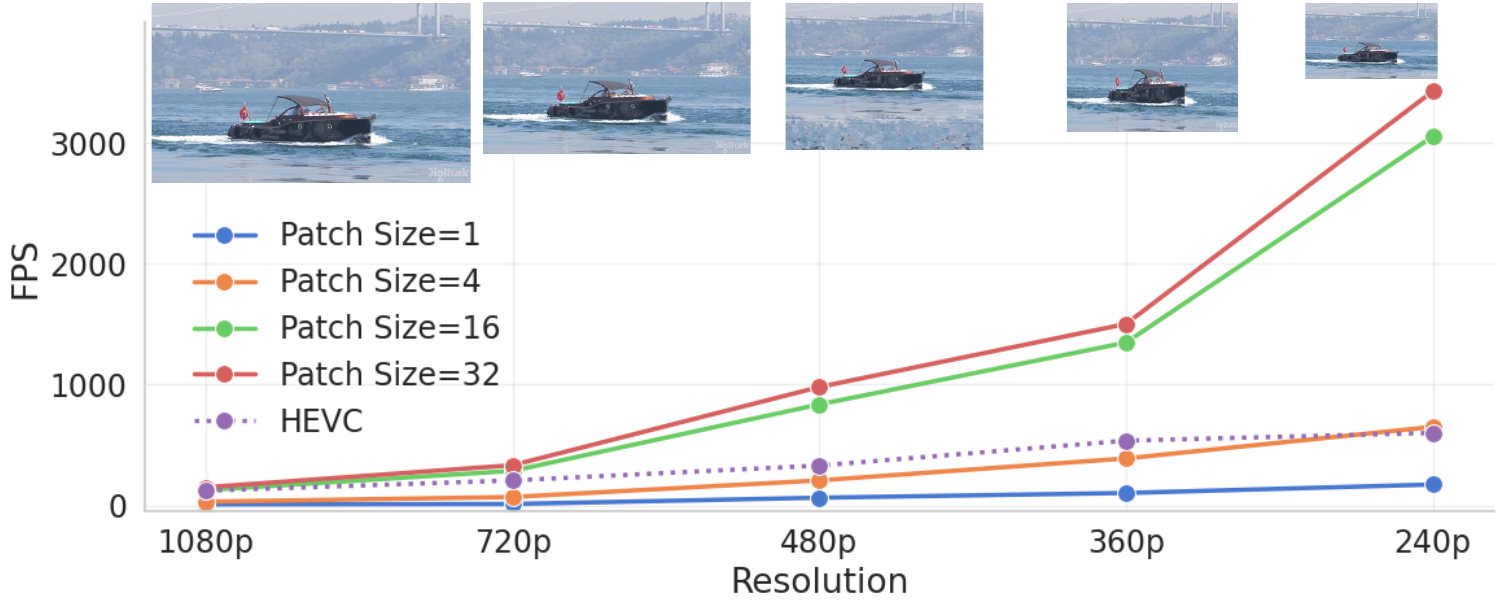}
    \caption{\small{With the same model, we can perform inference at any resolution, with speeds competitive or beating HEVC. We show sample frames for each resolution.}}
    \label{fig:any_res_inference}
  \end{minipage}
  \hfill
  \begin{minipage}{0.49\linewidth}
    \centering
    \includegraphics[width=\linewidth]{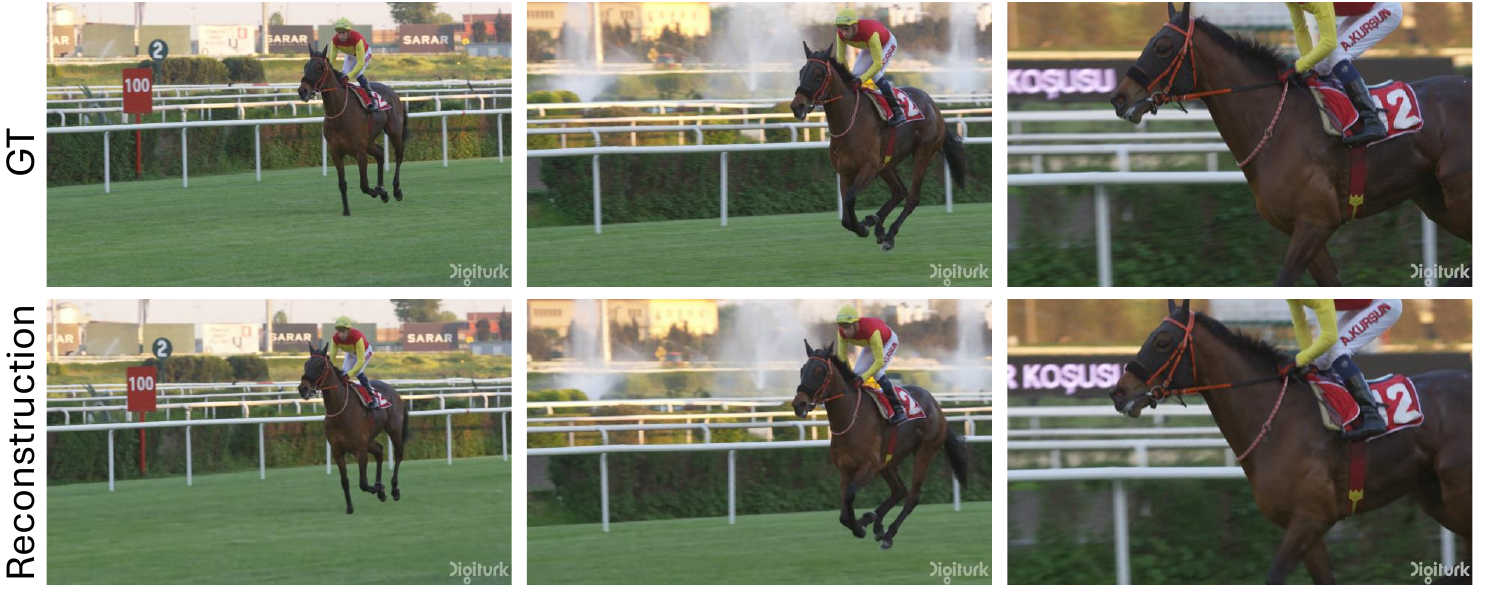}
    \caption{\small{We achieve high quality reconstruction and are able to reproduce even the finer details like water fountains and the hair on the horse. }}
    \label{fig:qualitative_reconstruction}
  \end{minipage}
\end{figure}
This dataset comprises seven high-quality videos, each featuring diverse scenes shot at 120fps over a duration of five seconds. While most videos contain 600 frames, the `shakendry' video is an exception with 300 frames, all at a resolution of 1080x1920. 
To assess the visual quality, we use standard metrics such as Peak Signal-to-Noise Ratio (PSNR) and Structural Similarity index (SSIM). 
We measure the storage efficiency of these methods using bits per pixel (BPP).
As mentioned earlier, we use feedforward MLPs for both the base network $f_{\theta}$ and hypernetworks $h_{l}$. 
The base network consists of 6 layers with layer size of 512 and each hypernetwork that modulates a selected layer has one hidden layer of size 128 with tanh non-linearity, followed by the output layer. 
In the case where we use patch centroids as inputs, we add a convolutional layer followed by a pixel-shuffle~\cite{Luo2022AnEE} for upsampling. 

We use hash-grids~\cite{mueller2022instant} for positional encoding due to their high quality reconstruction, although it should be noted we can use other schemes, such as Fourier features~\cite{tancik2020fourfeat} to exchange some quality for faster training (see Appendix).
We compare our method against NeRV~\cite{chen2021nerv} and NVP~\cite{kim2022scalable}, with each of them encoding a video per model, and the results are presented in Figure \ref{fig:rate_distortion}.
We observe that compression from our framework is comparable to baselines at similar bpp ranges, in addition to the other downstream benefits it offers. 

Due to our architecture, we are also able to operate in a novel paradigm,``\textbf{any-resolution inference}.''
Without changing the network architecture at all, we can decode the video at arbitrary smaller resolutions, as well as at higher resolutions (super-resolution) by leveraging the continuous resolution property of our hash grids and MLPs. 
We show our FPS decoding at various resolutions in Figure~\ref{fig:any_res_inference}, although it should be noted that HEVC, the standard codec we compare to, must encode separately for every resolution while we can store all in the same model. 
Figure~\ref{fig:qualitative_reconstruction} provides samples that showcase our method's fidelity.

\begin{table}[t]
\centering

    \begin{minipage}{0.475\linewidth}
        \caption{Interpolation Performance (PSNR), for different scale strides ($\alpha$).}
        \label{tab:interpolation_perf}
        \footnotesize
        \setlength\tabcolsep{5pt}
        \centering
        \renewcommand{\arraystretch}{1.5}
        \resizebox{1.0\columnwidth}{!}{%
        \begin{tabular}{@{}lccccc@{}}
        \toprule
        \textbf{Dataset} & \textbf{$\alpha$} & \textbf{NeRV} & \textbf{NIRVANA} & \textbf{NVP} & \textbf{Ours} \\
        \cmidrule{1-6}
        \multirow{3}{*}{\centering Bunny} & 2 & 15.92 & 19.14  & 20.10 & \textbf{33.17}  \\
        & 4 & 15.43 & 18.90  & 19.11 & \textbf{28.08} \\
        & 8 & 13.68 & 18.67  & 18.08 & \textbf{25.88} \\
        \midrule
        \multirow{3}{*}{\centering TaiChi} & 2 & 16.91 & 18.19  & 19.33 & \textbf{35.13}  \\
        & 4 & 17.14 & 17.71  & 18.52 & \textbf{31.84} \\
        & 8 & 15.72 & 16.21  & 17.7 & \textbf{27.72}\\
        \cmidrule{1-6}
        \end{tabular}}
    \end{minipage} 
    \hfill
    \begin{minipage}{0.475\linewidth} 
        \caption{Reconstruction and retrieval ablations of CLIP on MSR-VTT.}
        \label{tab:clip_lambda_ablation}
        \setlength\tabcolsep{5pt}
        \centering
        \renewcommand{\arraystretch}{1.5}
        \resizebox{1.0\columnwidth}{!}{%
        \begin{tabular}{@{}ccccc@{}}
        \toprule
        \multicolumn{1}{l}{} & \multicolumn{1}{c}{\textbf{Reconstruction}} & \multicolumn{3}{c}{\textbf{Retrieval (T2V)}} \\
        \cmidrule(lr){2-2} \cmidrule(lr){3-5}
        \textbf{CLIP $\lambda$} & \textbf{PSNR} & \textbf{R@1} & \textbf{R@5} & \textbf{R@10} \\ 
        \midrule
        0.0 & 30.03 & 0.1 & 0.3 & 0.8 \\
        1e-3 & 29.83 & 28.4 & 50.8 & 60.6 \\
        1e-2 & 29.46 & 30.2 & 52.4 & 61.0 \\
        1e-1 & 28.93 & 29.7 & 51.5 & 61.8 \\
        1.0 & 28.61 & 30.2 & 51.4 & 61.3 \\
        \bottomrule
        \end{tabular}
        }
    \end{minipage}  
\end{table}

\subsection{Video Interpolation}
\begin{figure}[t]
    \centering
    \includegraphics[width=\linewidth]{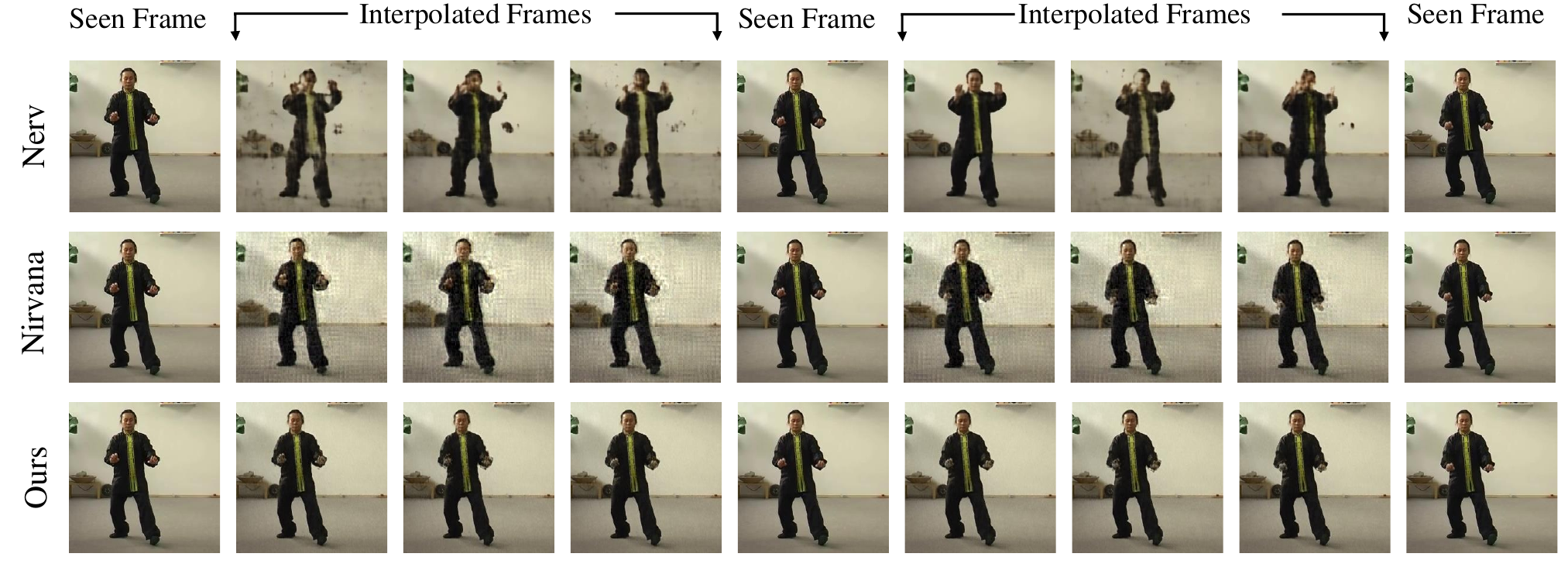}
    \caption{\small{We compare interpolation with Latent-INR to NVP and NIRVANA. We find that our method has less artifacts and smoother motion in the interpolated frames.}}
    \label{fig:interp_fig}
\end{figure}

In our framework, we can interpolate in the latent space to generate valid interpolated frame outputs.
We conduct experiments on two datasets: the ``big buck bunny sequence" and a selection of ten videos from the Taichi test set. 
Frames are held out at a scale stride $\alpha$ during encoding. 
During testing, we interpolate the resulting latents on the held out frames and evaluate their performance.

We use the same INR models utilized for compression as our baselines, with a reduction in network layer size and modulating mask rank.
While NeRV \cite{chen2021nerv} and NVP \cite{kim2022scalable} interpolate time positions used as input, NIRVANA interpolates the weights. 
In Table \ref{tab:interpolation_perf}, we observe that while other INR methods fail to produce perceptual frames at scale of 2, our model can give reasonable interpolations even at a scale of 8. 
We confirm this qualitatively also, by inspecting interpolated frames such as those shown in Figure~\ref{fig:interp_fig}.
Our outputs have noticeably fewer artifacts, and while imperfect, handle the motion better.
Compared to other video INR methods, our approach of using learnt latents facilitates the model to have an internal representation of the video content. 

\subsection{Downstream Tasks}
\label{subsec:exp_retrieval}

\begin{table}[t]
\centering
    \begin{minipage}{0.485\linewidth} 
        \caption{\small{Class and segment retrieval. Our method often exceeds CLIP performance.}}
        \label{tab:retrieval_perf}
        \setlength\tabcolsep{5pt}
        \centering
        \renewcommand{\arraystretch}{1.5}
        \resizebox{1.0\columnwidth}{!}{%
        \begin{tabular}{@{}cccccccc@{}}
        \toprule
        \multicolumn{1}{l}{}& & \multicolumn{3}{c}{\textbf{Class Level}} & \multicolumn{3}{c}{\textbf{Segment Level}} \\
        \cmidrule(lr){3-5} \cmidrule(lr){6-8}
        \multicolumn{1}{l}{\textbf{Dataset}}                     &     \textbf{Method}            & \multicolumn{1}{c}{R@1} & \multicolumn{1}{c}{R@5} & \multicolumn{1}{c}{R@10} & \multicolumn{1}{c}{R@1} & \multicolumn{1}{c}{R@5} &\multicolumn{1}{c}{R@10} \\ \toprule
        
        \multirow{2}{*}{\centering COIN} & CLIP & 31.60 & 44.70  & \textbf{50.70} & \textbf{6.60} & 13.10 & 16.50 \\
        & Ours & \textbf{34.40} & \textbf{45.10}  & 50.50 & 6.40 & \textbf{13.30} & \textbf{17.00}\\
        \midrule
        \multirow{2}{*}{\centering HowTo100m*} & CLIP & \textbf{31.58} & 36.84  & 47.37 & 21.13 & 37.32 & 40.85 \\
        & Ours & 31.58 & \textbf{42.11}  & \textbf{47.36} & \textbf{23.24} & \textbf{43.67} & \textbf{48.60}\\
        \bottomrule
        \end{tabular}
        }
    \end{minipage} 
    \hfill
    \begin{minipage}{0.485\linewidth} 
        \caption{\small{Whole video retrieval. Our method matches CLIP performance.}}
        \label{tab:retrieval_perf_whole_video}
        \setlength\tabcolsep{5pt}
        \centering
        \renewcommand{\arraystretch}{1.5}
        \resizebox{1.0\columnwidth}{!}{%
        \begin{tabular}{@{}cccccccc@{}}
        \toprule
        \multicolumn{1}{l}{}&  & \multicolumn{3}{c}{\textbf{Text to Video}} & \multicolumn{3}{c}{\textbf{Video to Text}} \\
        \cmidrule(lr){3-5} \cmidrule(lr){6-8}
        \multicolumn{1}{l}{\textbf{Dataset}}                     &    \textbf{Method}       & \multicolumn{1}{c}{R@1} & \multicolumn{1}{c}{R@5} & \multicolumn{1}{c}{R@10} & \multicolumn{1}{c}{R@1} & \multicolumn{1}{c}{R@5} &\multicolumn{1}{c}{R@10} \\ \toprule
        
        \multirow{2}{*}{\centering MSR-VTT} & CLIP  & 30.10 & 51.50 & \textbf{61.50} & 24.70 & 49.30 & \textbf{61.90} \\
        & Ours & \textbf{30.20} & \textbf{52.40} & 61.10 & \textbf{25.40} & \textbf{49.90} & 61.70 \\
        \midrule
        \multirow{2}{*}{\centering ActivityNet*} & CLIP & 38.4 & \textbf{74.8} & \textbf{86.6} & \textbf{36.2} & \textbf{73.6} & \textbf{84.8} \\
        & Ours & \textbf{38.5} & 73.9 & 86.4 & 36.1 & 73.5 & 84.7 \\
        \bottomrule
        \end{tabular}
        }
    \end{minipage}  
\end{table}

\begin{figure}[t]
    \centering
    \includegraphics[width=\linewidth]{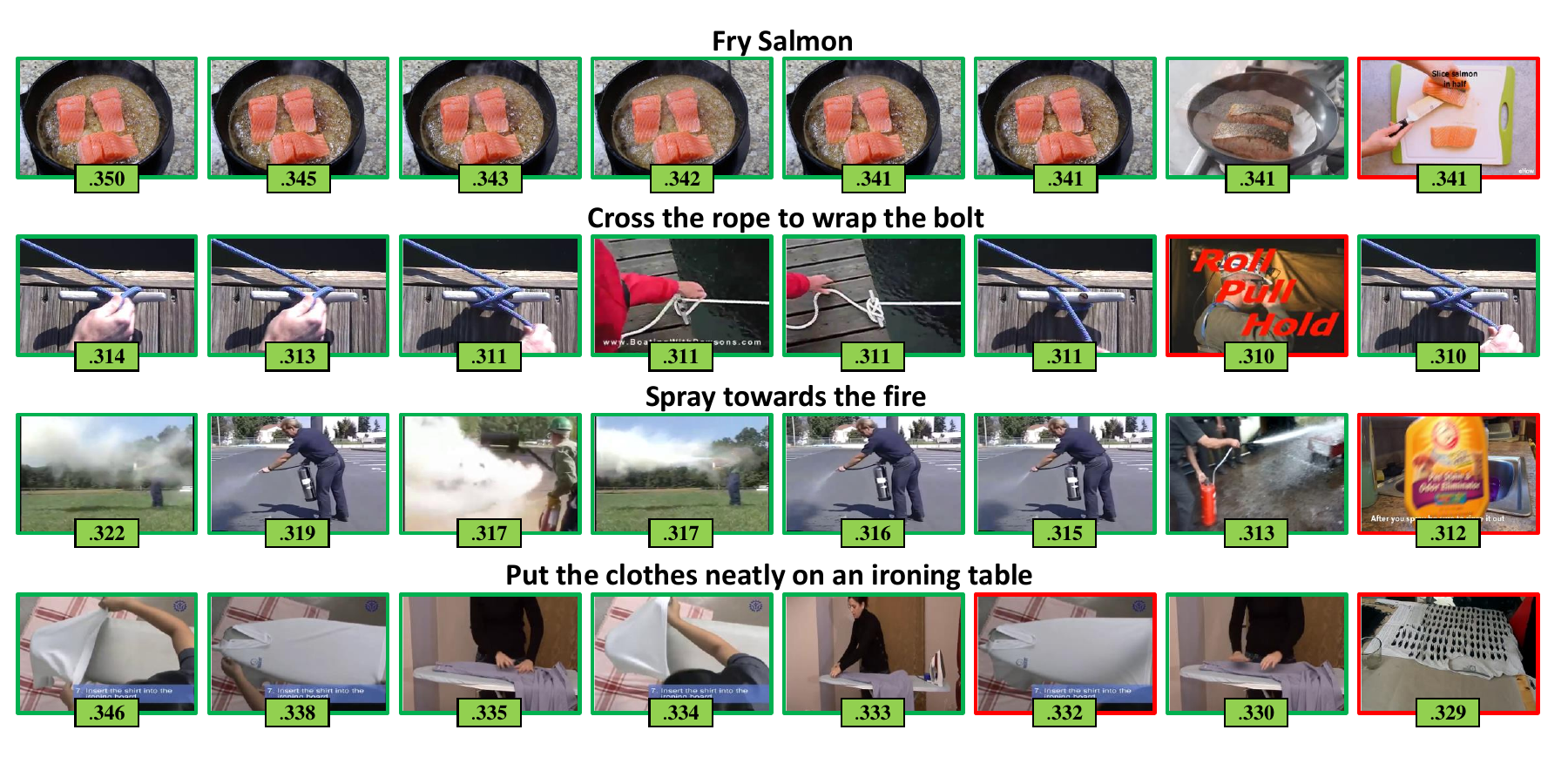}
    \caption{\small{Nearest Neighbours for segment-level matching of sample queries from COIN validation set. The green boxes denote the true positives and the red ones are false positives. We show the inner product similarity between the image and the corresponding query inside the green boxes at the bottom of each image.}}
    \label{fig:retrieval}
\end{figure}

\textbf{Retrieval}

To showcase the flexibility of our latents, we align them with CLIP and evaluate their performance on standard retrieval tasks. 
We utilize the validation set of COIN dataset \cite{tang2019coin} and a subset of Howto100m dataset to evaluate performance. 
We first encode each video in our split using our Latent-INR framework with a loss that encourages the latents to be closer to the CLIP-Image embeddings of the frames, in addition to the standard reconstruction loss. 
We consider two distinct problems --
retrieval of the correct class across all videos and retrieval of the correct segment within a video. These two use cases cover both ends of the spectrum, from localizing an event in a given video to searching for similar events across videos.
We utilize the standard recall at K, where we have selected $k\in [1,5, 10]$ to evaluate the efficacy of our method.
The results are presented in Table~\ref{tab:retrieval_perf}. We can see that our method matches CLIP in its retrieval performance and even exceeds it in some cases. The qualitative results are presented in Figure \ref{fig:retrieval}, where we visualize the top 5 nearest neighbours of the text query that map to trained latents across all videos. Further results can be found in the supplementary.
We even find that our method can perform whole-video retrieval on MSR-VTT~\cite{Xu_2016_CVPR} and a custom 1,000 video sample from the ActivityNet Captions~\cite{krishna2017densecaptioning} `val-1' split.
We average-pool both our features and CLIP features (similar to \cite{buch2022revisiting}) and use CLIP features computed on video captions.
In Table~\ref{tab:retrieval_perf_whole_video} we find that our retrieval is quite competitive to retrieval using the CLIP features themselves, showing that the learnt latents have similarly good averaging and summarizing properties even over longer (180 seconds) videos, as well as alignment even to the paragraph-length captions used in ActivityNet.

\begin{figure}[t]
    \centering
    \includegraphics[width=\textwidth]{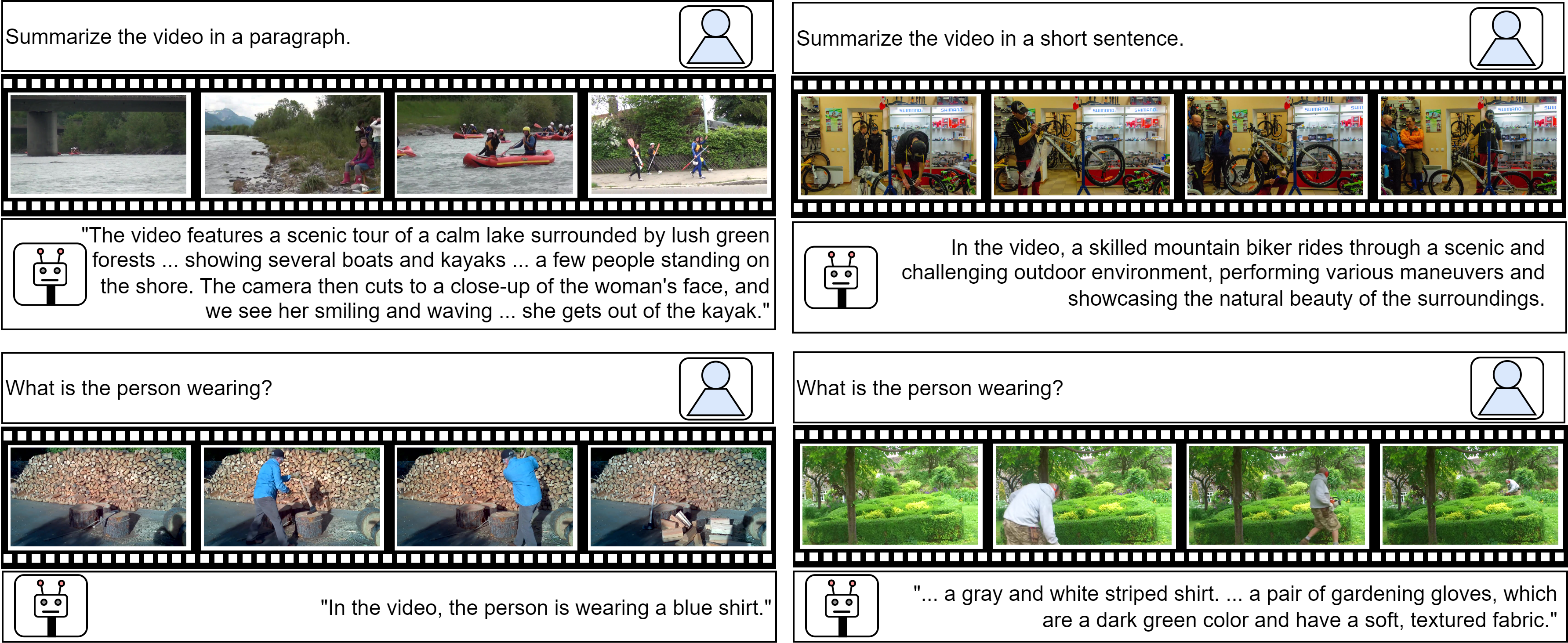}
    \caption{Latent-INR LLM. We show results for aligning our learned latents to a VideoLlama model, which allows for interactive chat. We show successes (left) and failures (right) for summarization (top) and question answering (bottom).}
    \label{fig:latent_inr_llm}
\end{figure}

\noindent\textbf{Video-based Chat}

We evaluate the performance of our trained latents, when aligned to intermediate VideoLlama features.
This alignment enables access to the full scope of text chat with video understanding.
We show a sample of such results, in the form of text and video prompts with text response, in Figure~\ref{fig:latent_inr_llm}.
These results show the LLM is able to understand video inputs when provided in the form of INR latents rather than raw video tokens.
While not perfect, we infer the majority of the shortcomings of this system are primarily the fault of the LLM we align to.
Furthermore, on the basis of our success in aligning with CLIP and now VideoLlama, we believe our latents can be aligned to any representation.
So, for more powerful chat, one simply needs to align to a more powerful chatbot.
We thus provide these results two purposes.
First, we show our model's capability to power efficient open-ended captioning and question answering, while still retaining reconstruction capabilities.
Second, we point to the immense potential of our model (or a similar paradigm) to continue to be leveraged with such models as they expand in their size and performance.

\subsection{Visualizing Trained Latents}

\begin{figure}[t]
    \centering
    \includegraphics[width=\textwidth]{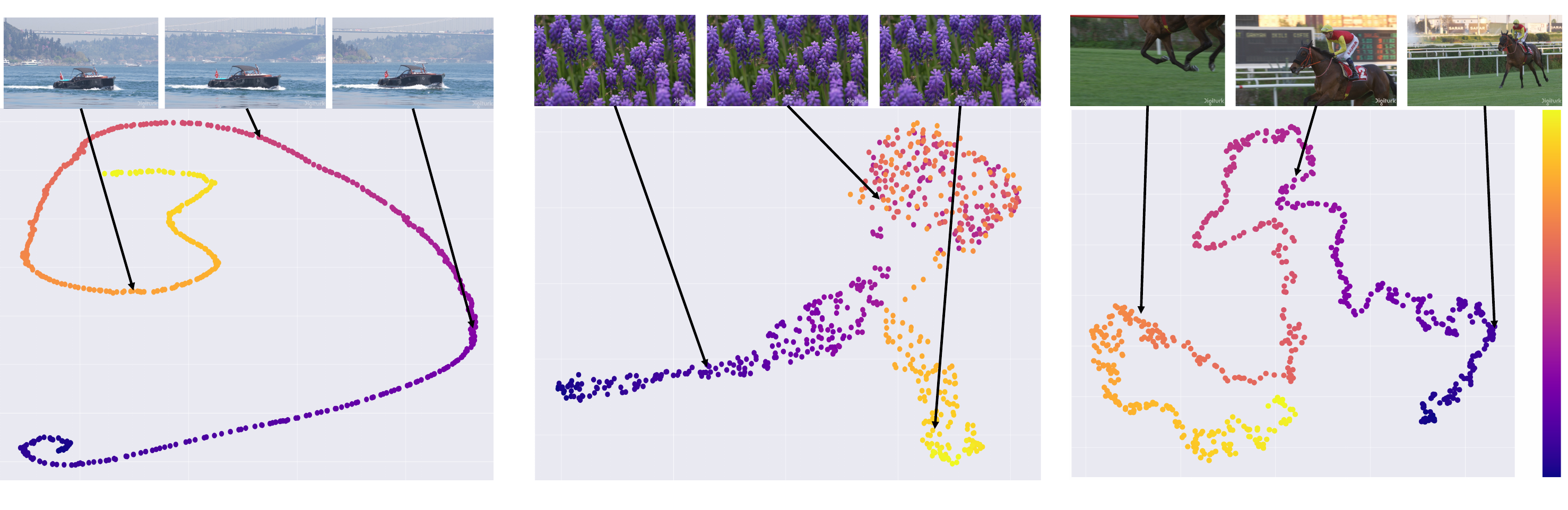}
    \caption{\small{We visualize the trained latents $Z_{t}$ projected to 2D using UMAP. We show that the trained latents from our framework capture meaningful semantics of the underlying data. Latents for Bosphore (left), Honeybee (middle) and Jockey (right) from UVG dataset. Dark to Light color indicates frame numbers ranging from 0 to 600.}}
    \label{fig:latent_viz}
\end{figure}

The trained latents, representing the modulated frames, offer intriguing insights when visualized in a reduced dimensional space. 
Utilizing Uniform Manifold Approximation and Projection (UMAP)\cite{mcinnes2020umap} we project the embeddings $Z_{t}$ into a 2D space, allowing for an intuitive interpretation of their relationships. 
In Figure~\ref{fig:latent_viz}, we plot the UMAP for three distinct videos from the UVG dataset: `Bosphore,' `Honeybee,' and `Jockey,' each offering unique characteristics for examination.

`Bosphore', characterized by its slow-moving object and relatively static foreground, exhibits a smooth latent trajectory in the 2D space. 
This smoothness reflects the minimal variance in frame content, suggesting that our method effectively captures the subtle dynamics of the scene. 
In contrast, the `Honeybee' video, with its repetitive frames, results in latents that cluster tightly together, signifying our model's ability to recognize and encode repetitive patterns efficiently. 
The most dynamic of the three, `Jockey', presents a more complex scenario with rapid changes in both the foreground and background. 
Here, the latents form clusters around similar scenes, yet maintain a discernible trajectory through the 2D space. These visualizations illustrate the semantic richness embedded within the latents obtained from our framework even when trained only for compression.


\section{Ablation Studies}

\begin{figure*}[ht]
    \centering
    \begin{subfigure}[b]{0.45\textwidth}
        \includegraphics[width=\textwidth]{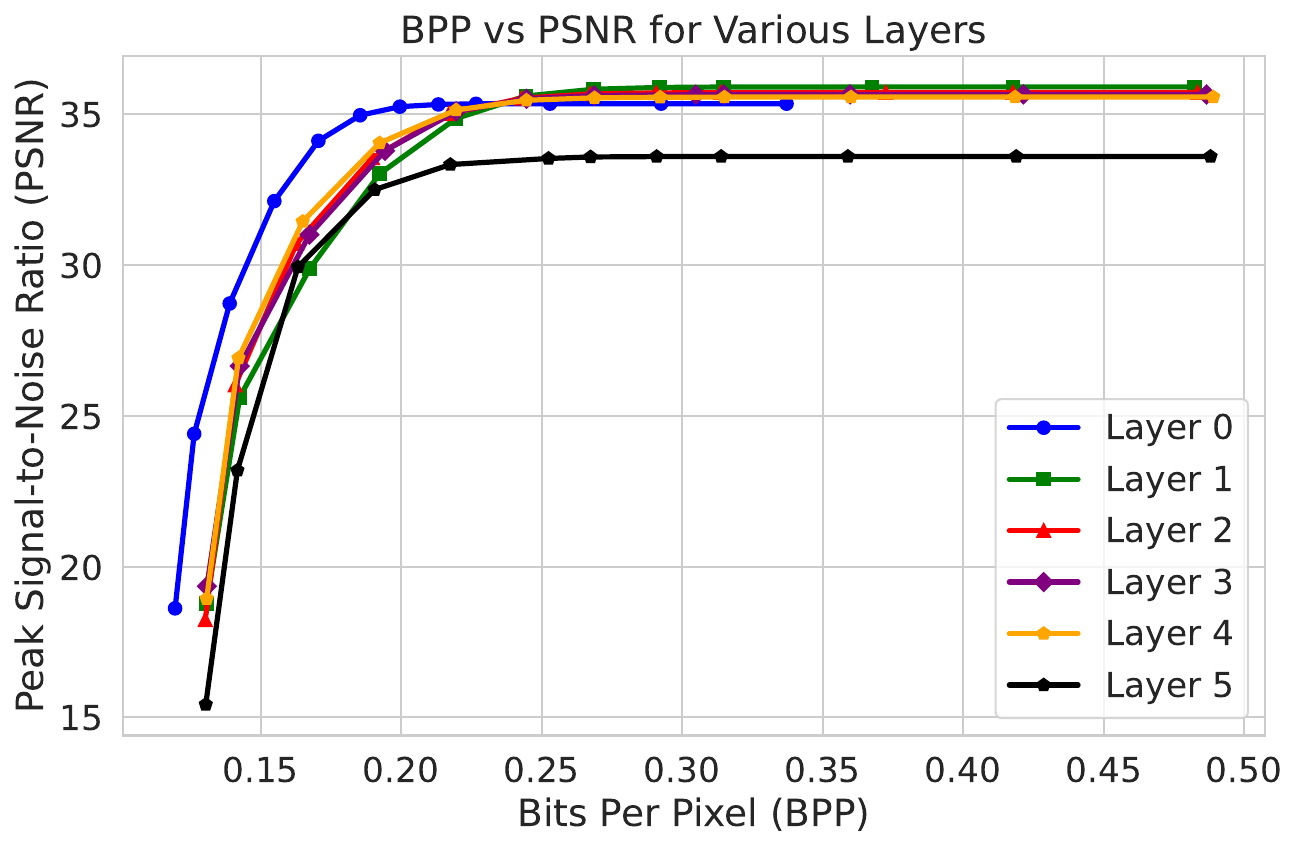}
        \label{fig:sub1}
    \end{subfigure}
    \begin{subfigure}[b]{0.45\textwidth}
        \includegraphics[width=\textwidth]{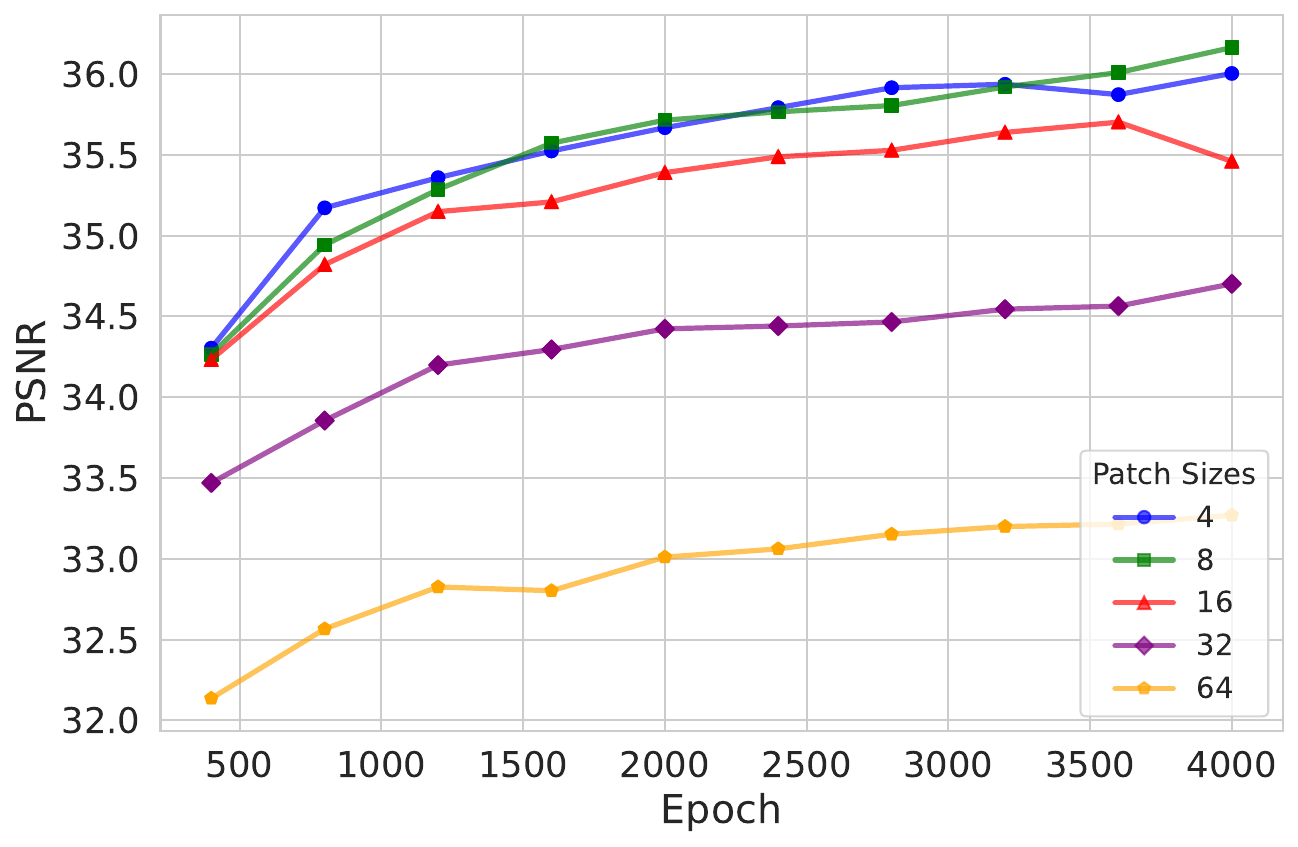}
        \label{fig:sub3}
    \end{subfigure}
    \caption{Ablations to study the effect of layer modulations in the hypernetwork and the effect of patch size on reconstruction quality (PSNR).}
    \label{fig:ablation}
\end{figure*}


\noindent\textbf{CLIP $\lambda$.} 
We investigate the impact of the large model alignment weighting term on both reconstruction and retrieval for MSR-VTT.
In Table~\ref{tab:clip_lambda_ablation}, we find that PSNR decreases slightly as $\lambda$ increases.
However, the retrieval performance seems to saturate at $\lambda = 0.01$.
So, we suggest not tuning the $\lambda$ too high for any application, given the diminishing returns.

\noindent\textbf{Layer Modulations.} 
In our approach, we have separate hypernetworks that modulated the selected layers. 
To evalute the importance of each, we design an experiment where they are modulated in isolation.
We use the same setup as the compression experiments with the modulating mask rank fixed at 20 for all models. 
In Figure~\ref{fig:ablation}, we can clearly see that the first few layers have a significant impact on the encoding performance. 
This matches the observations from \cite{kim2023generalizable} about the out sized impact of first few layers while modulating INRs. 


\noindent\textbf{Patch Size.} 
Scaling to higher-resolution videos can be memory-intensive. 
This is particularly true when employing memory-demanding positional encoding schemes such as hash-grids~\cite{mueller2022instant}. 
To investigate this aspect further, we experiment with models that process centroids of fixed-size patches, directly predicting the corresponding frame patches, to save memory. 
From Figure \ref{fig:ablation}, performance is consistent for smaller patch sizes, but drops off sharply for higher patch sizes. 

\section{Conclusion}

\noindent\textbf{Limitations.} 
Our latents are somewhat restricted by the quality of the embeddings they are aligned to.
Additionally, more work is still required to match standard codecs in terms of storage and encoding time, in spite of impressive gains in terms of quality and decoding speed. Future work could both improve the compression, and leverage more powerful vision models.

\noindent\textbf{Broader Impacts.}
Our method for simultaneously compressing and learning useful features for recognition could reduce the need to decode videos for these tasks and thus save computational resources, cutting costs and helping the environment.
However, work that advances performance for compression and recogntion also has applications in surveillance and warfare.

In this work, we propose a new framework, Latent-INR, where we decouple the temporal aspect from the spatial into a dictionary of learnable latents. 
These auto-decoder based learnable latents modulate the layers of the base INR network via low-rank modulation using hypernetworks. 
Latent-INR is not only well-suited to video compression, but the resulting latents learn an internal representation of the data they encode that lends itself to SOTA interpolation for video INRs. 
Additionally, we also augment these latents by training them to be aligned with CLIP and VideoLlama, which allows us to bring the power of foundational models to compressed representations, and perform retrieval and chat-based applications like captioning and question answering. 
Our work thus opens up new possibilities of research in the implicit neural space where downstream tasks can be performed by these model without the need for decoding.

\noindent\textbf{Acknowledgements}
This work was partially supported by NSF CAREER Award (\#2238769) to AS. The U.S. Government is authorized to reproduce and distribute reprints for Governmental purposes notwithstanding any copyright annotation thereon. The views and conclusions contained herein are those of the authors and should not be interpreted as necessarily representing the official policies or endorsements, either expressed or implied, of NSF or the U.S. Government.
%
%

\clearpage
\setcounter{page}{1}
\appendix

\section{Network Architecture }

\label{sec:arch_details}
\textbf{Base Network:} We use an MLP with 10 layers, width of 512 and ReLU non-linearity as our base network $f_{\theta}$. \\
\textbf{Hypernetwork:} All hypernetworks $h^{l}$ used to modulate a layer $l$ of the base network have 3 layers with a hidden dimension of 512 and tanh as non-linearity. Unless specified, we only modulate the first hidden layer of the base network. \\
\textbf{Latents:} Each latent $Z_{t}$ corresponding to a frame has a dimension of 512 and is initialized to be standard Gaussian before training. We set our learning rate as 5e-4 and used the standard Adam optimizer without any weight decay. 

\section{Compression}
\subsection{Fourier Features}
We use the multiresolution hash grid for positional encoding in all our models. In table \ref{tab:fourier} we show results for full coordinate resolution using fourier features for positional encoding. Due to lack of a hash grid, the resulting models train upto 30\% faster, but at the cost of inferior reconstruction.

\subsection{Quantization}
Instead of quantizing all components equally, we notice that retaining the latents and the base network at full precision provides better reconstruction at negligible additional storage.

\subsection{Effect of latent dimension}
\begin{figure}[ht]
    \centering
    \begin{subfigure}[b]{0.45\textwidth}
        \includegraphics[width=\textwidth]{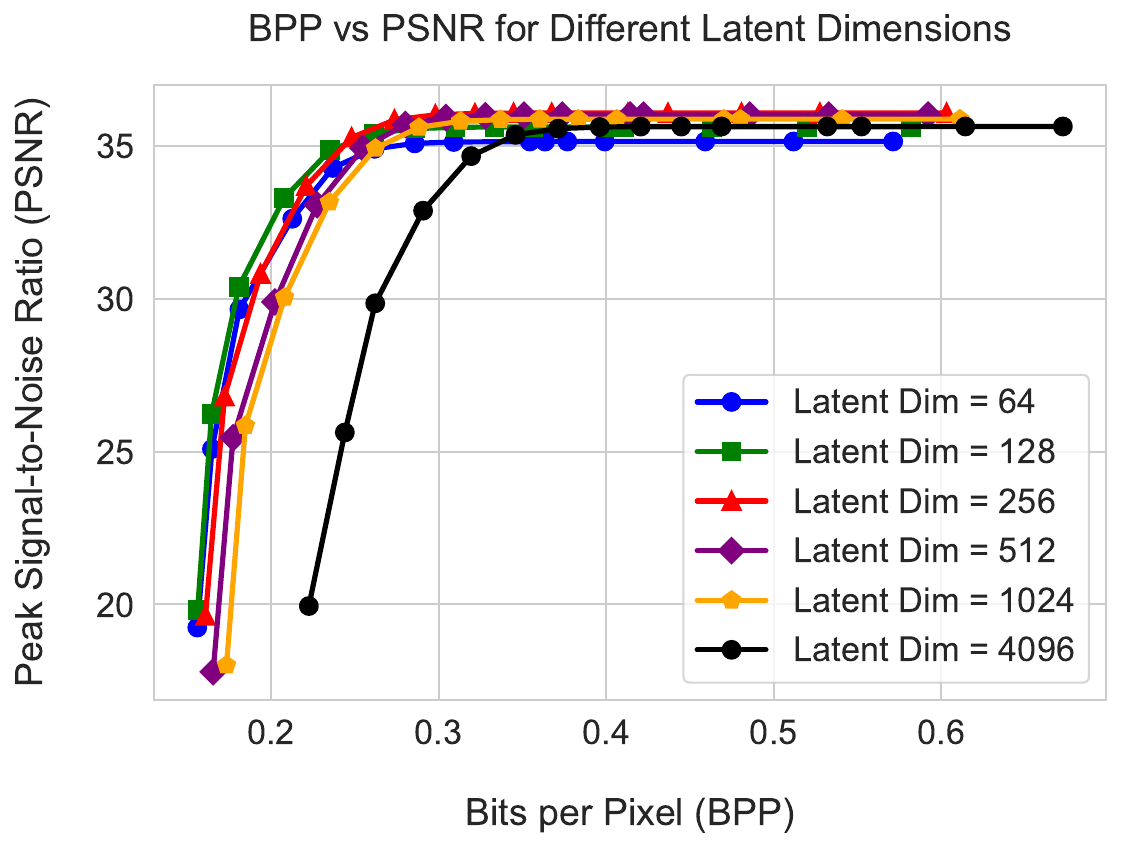}
    \end{subfigure}
    \begin{subfigure}[b]{0.45\textwidth}
        \includegraphics[width=\textwidth]{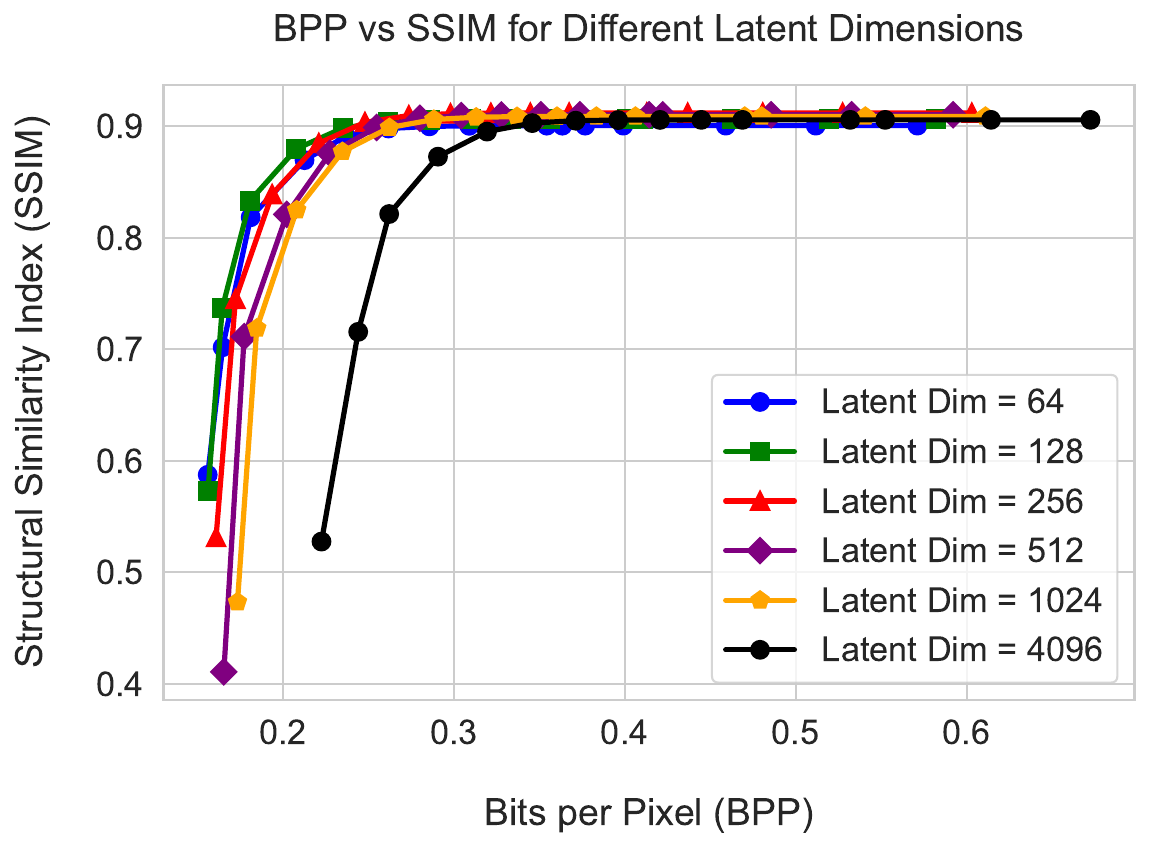}
    \end{subfigure}
    \caption{Effect of varying latent dimension across different bitrates.}
    \label{fig:vary_latent}
\end{figure}

To study the effect of latent dimension on compression, we train models by varying it and encode the ``bosphore'' video from UVG dataset. The results are presented in Figure \ref{fig:vary_latent}. We notice that there is positive gains till dimension 512 and diminishing returns thereafter. Hence we choose that as our default latent size in all our experiments. 

\section{ Video Retrieval}
We perform two retrieval tasks on the COIN dataset\cite{tang2019coin} - \textit{class-level}, and \textit{segment-level}. In both settings, we use the standard val set as the database. For \textit{class-level}, we use the distinct video-level task names in COIN as our query set. For \textit{segment-level}, we use the set of distinct clip-level captions in COIN as our query set. We get the CLIP ViT-B/32 text embeddings of each of these captions, and these become our query vectors. For database vectors, we use the per-frame learned latents for each video in the database. For comparison with CLIP, we replace these database vectors with the CLIP ViT-B/32 image embeddings for each frame.  For \textit{class-level} retrieval, we consider a result frame a positive match if it belongs to a video with the same class label as the queried caption. On the other hand, for \textit{segment-level} retrieval, we consider a result frame a positive match only if it belongs to a segment with the same caption as the query. Further, this search is done over all videos. We use FAISS\cite{johnson2019billion} as our retrieval implementation and use \textit{cosine similarity} as the distance metric.

We perform whole-level video retrieval as described in the main paper. 
For text, we use CLIP to compute a feature for the paragraph caption.
For the video, we compute a per-frame feature for CLIP, or use the learnt latents from Latent-INR.
For a single video feature, we then average these per-frame features.
We normalize all features, and perform retrieval by finding the closest embeddings using dot product similarity.
Both text-to-video and video-to-text are performed in the same manner, the only difference being which features are used as query and key.

\begin{figure} [t]
    \centering
    \includegraphics[width=\linewidth]{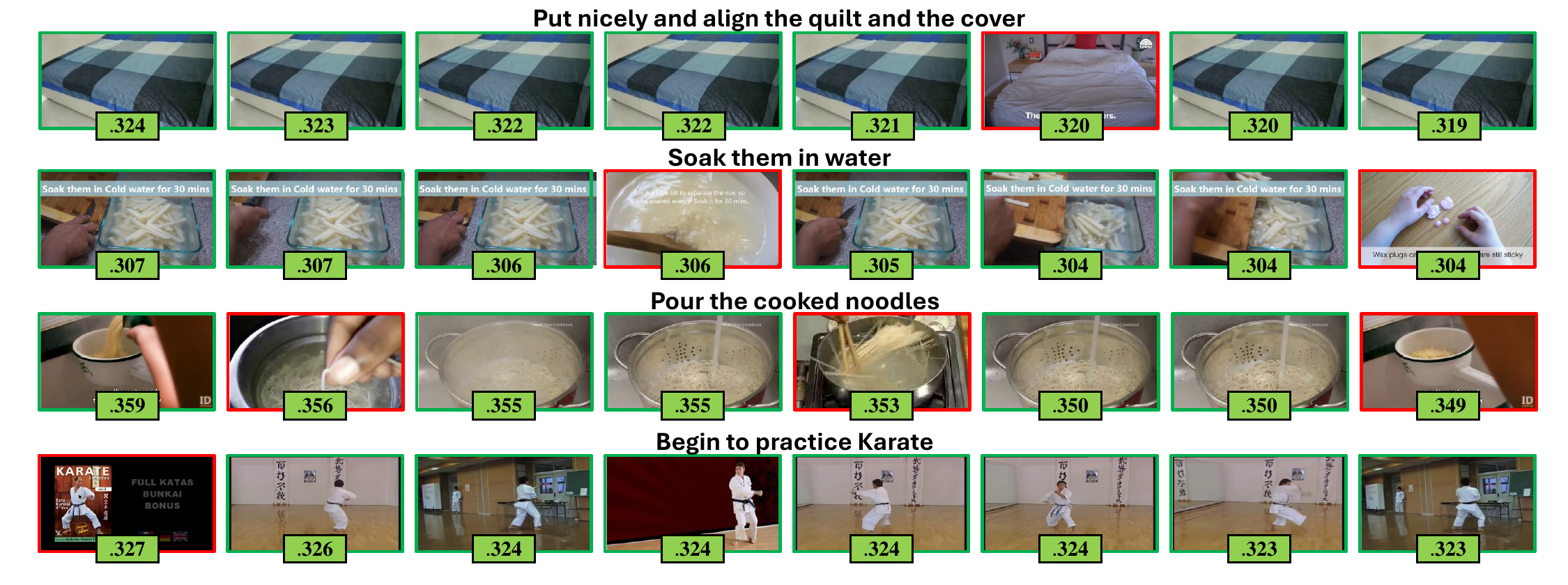}
    \caption{Nearest Neighbours for segment-level matching of sample queries from COIN
validation set. The green boxes denote the true positives and the red ones are false pos-
itives. We show the inner product similarity between the image and the corresponding
query inside the green boxes at the bottom of each image}
    \label{fig:retfig_supp}
\end{figure}

Fig.\ref{fig:retfig_supp}, shows the retrieval results on the COIN data in the \textit{segment-level} setting. It can be seen that a majority of failure cases could be attributed to visual similarity across different tasks when seen at an individual frame level.

\section{Video Chat}
\begin{figure}[t]
    \centering
    \includegraphics[width=\textwidth]{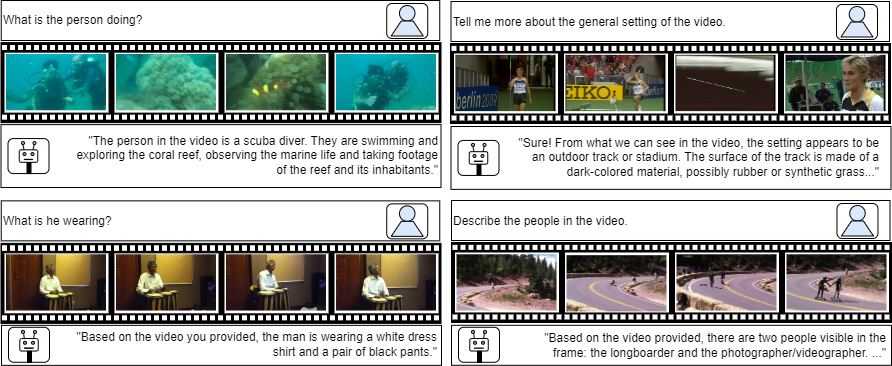}
    \caption{Additional results for Latent-INR interface with Video-LLM.}
    \label{fig:latent_inr_llm_app}
\end{figure}

We interface our latents with learned features from Video-Llama \cite{damonlpsg2023videollama} to enable interactive chat with the compressed videos. In \cite{damonlpsg2023videollama}, the $N$ video frames are passed through a ViT based visual encoder to extract features of size $k \times d$ per frame. These are then passed through a Query Former \cite{10.5555/3618408.3619222} to obtain a unified video representation of size $k_{v} \times d_{v}$. This tensor is then passed to a trainable MLP layer before aligning with an LLM of our choice (LLama-2 \cite{touvron2023llama} in our models). \\
We align our latents $Z$ with these per-video features of size $k_{v} \times d_{v}$ using a linear projection layer which is trained end to end. The loss function is slightly modified to incorporate a cosine similarity loss between the terms.

\begin{equation}
    \begin{aligned}
        L = L_{\text{MSE}} + \lambda \cdot  L_{cos}(F_{t},F_{t}^{\text{V-LLM}})  
    \end{aligned}
\end{equation}
where $F_{t}$ is the predicted feature and $F_{t}^{\text{V-LLM}}$ is the corresponding Video-LLama extracted features.
We show additional results of the interactive chat in Figure \ref{fig:latent_inr_llm_app}.


\begin{table}[t]
\caption{Fourier Feature Models}
\label{tab:fourier}
\centering
\renewcommand{\arraystretch}{1.2} 
\begin{tabular}{@{}ccccc@{}}
\toprule
\textbf{Method} & \textbf{PSNR} & \textbf{BPP} \\ 
\midrule
Ours- Fourier - \textit{S}  & 31.99 & 0.31  \\ 
\hline 
Ours- Fourier - \textit{M}  & 33.69 & 0.62  \\ 
\hline 
Ours- Fourier - \textit{L}   & 33.19 & 0.84 \\ 
\bottomrule
\end{tabular}
\end{table}



\section{Video-wise results}

\begin{figure}[ht]
    \centering
    \begin{subfigure}[b]{0.45\textwidth}
        \includegraphics[width=\textwidth]{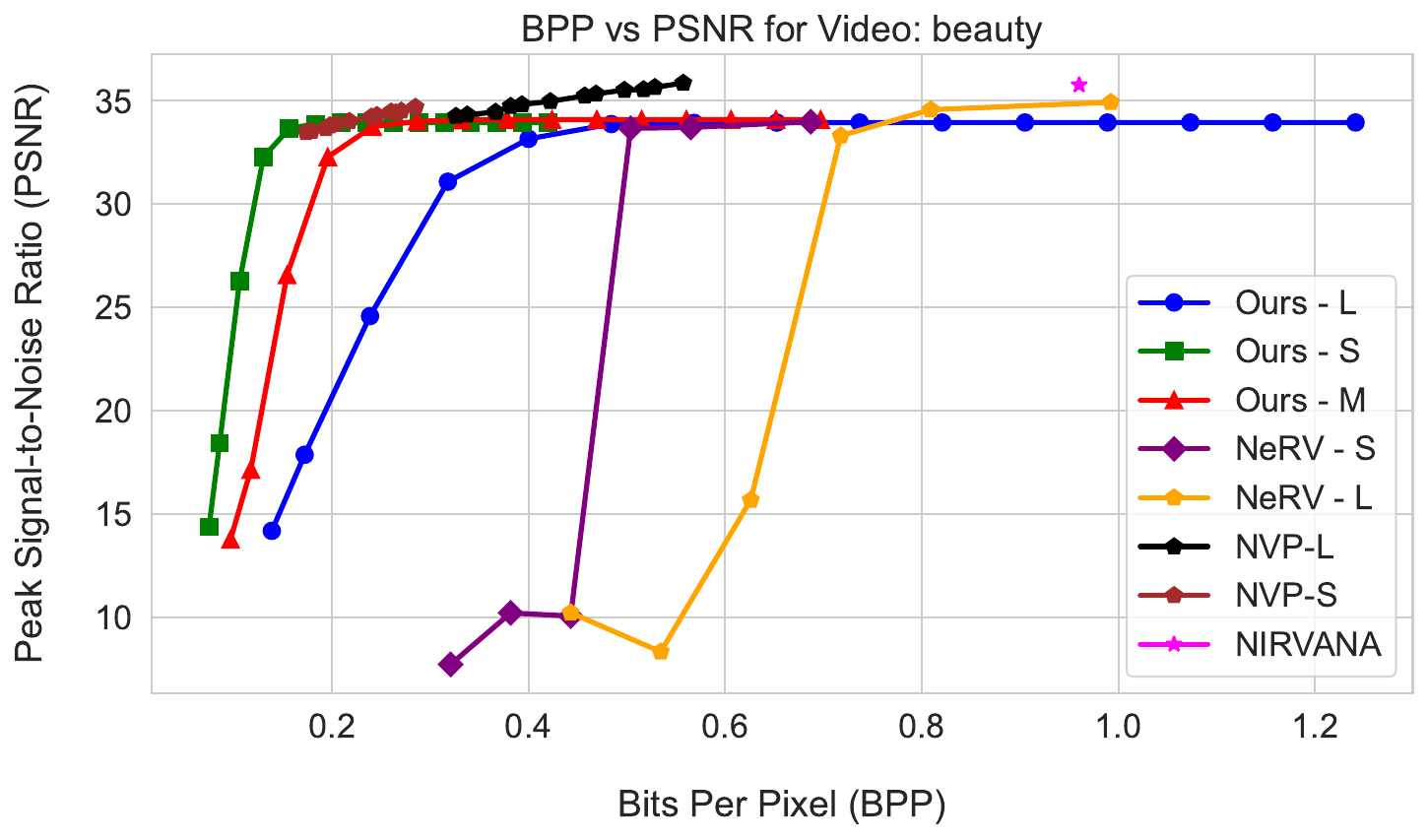}
    \end{subfigure}
    \begin{subfigure}[b]{0.45\textwidth}
        \includegraphics[width=\textwidth]{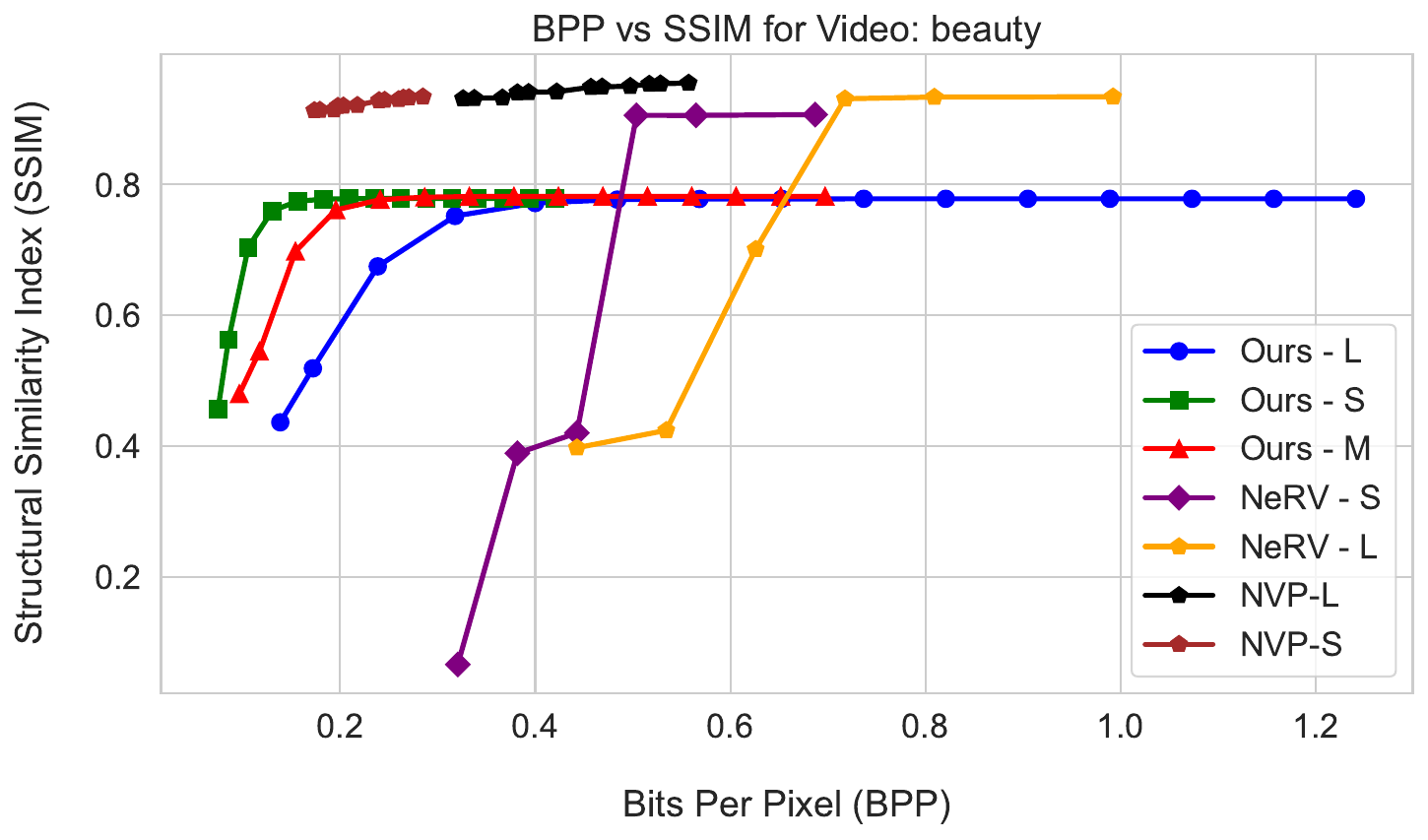}
    \end{subfigure}
    \caption{BPP vs. PSNR, SSIM for beauty.}
    \label{fig:beauty}
\end{figure}

\begin{figure}[ht]
    \centering
    \begin{subfigure}[b]{0.45\textwidth}
        \includegraphics[width=\textwidth]{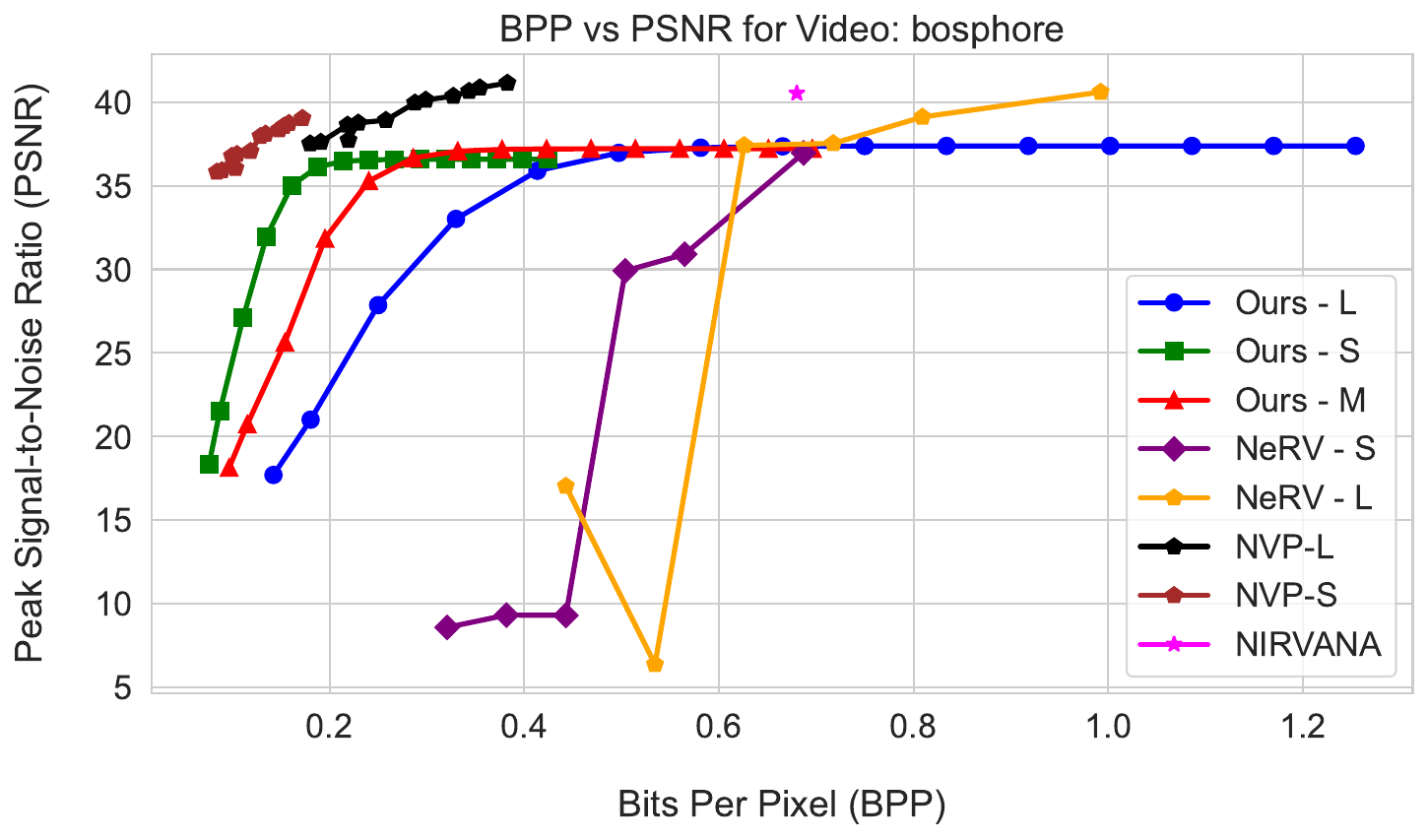}
    \end{subfigure}
    \begin{subfigure}[b]{0.45\textwidth}
        \includegraphics[width=\textwidth]{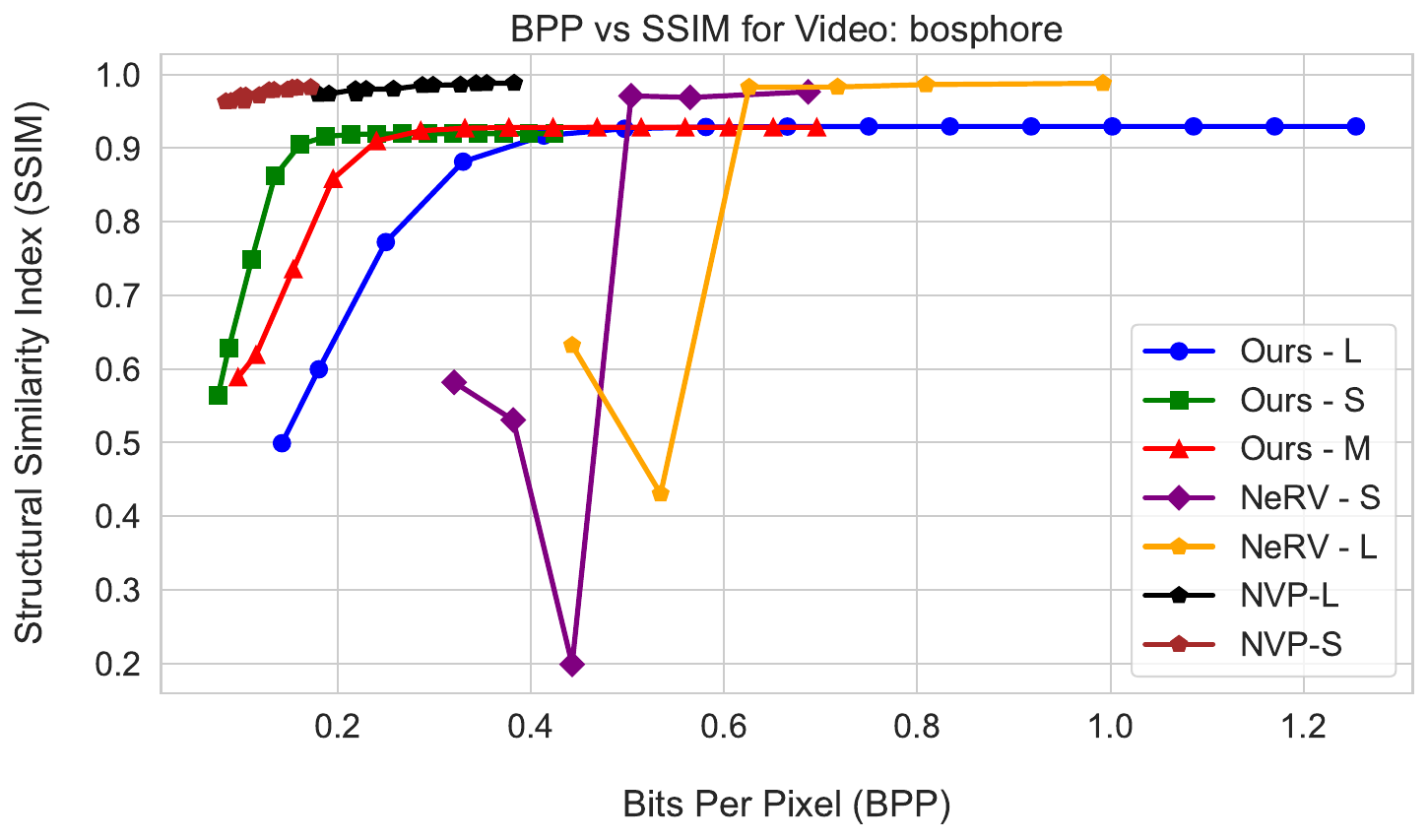}
    \end{subfigure}
    \caption{BPP vs. PSNR, SSIM for bosphore.}
    \label{fig:bosphore}
\end{figure}

\begin{figure}[ht]
    \centering
    \begin{subfigure}[b]{0.45\textwidth}
        \includegraphics[width=\textwidth]{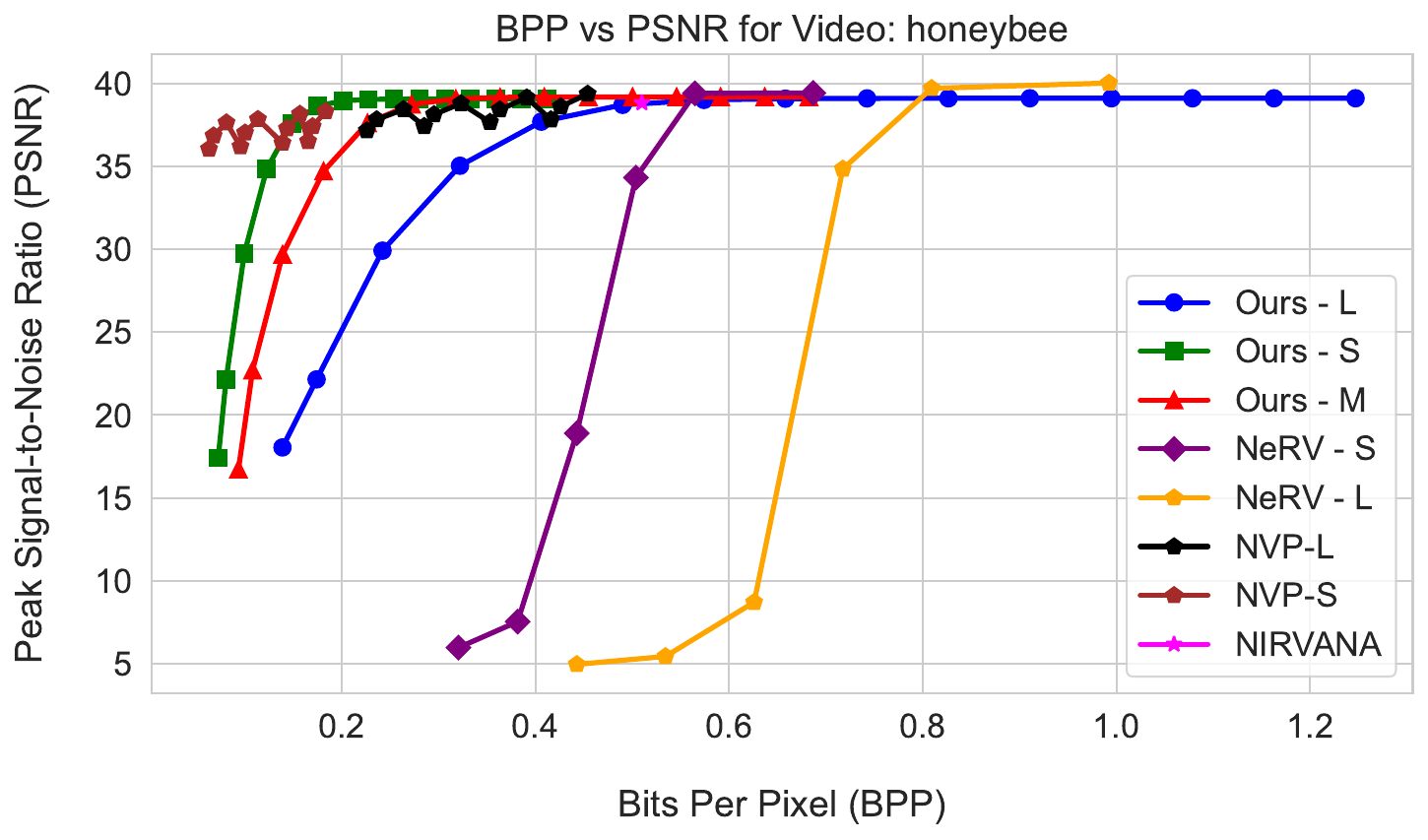}
    \end{subfigure}
    \begin{subfigure}[b]{0.45\textwidth}
        \includegraphics[width=\textwidth]{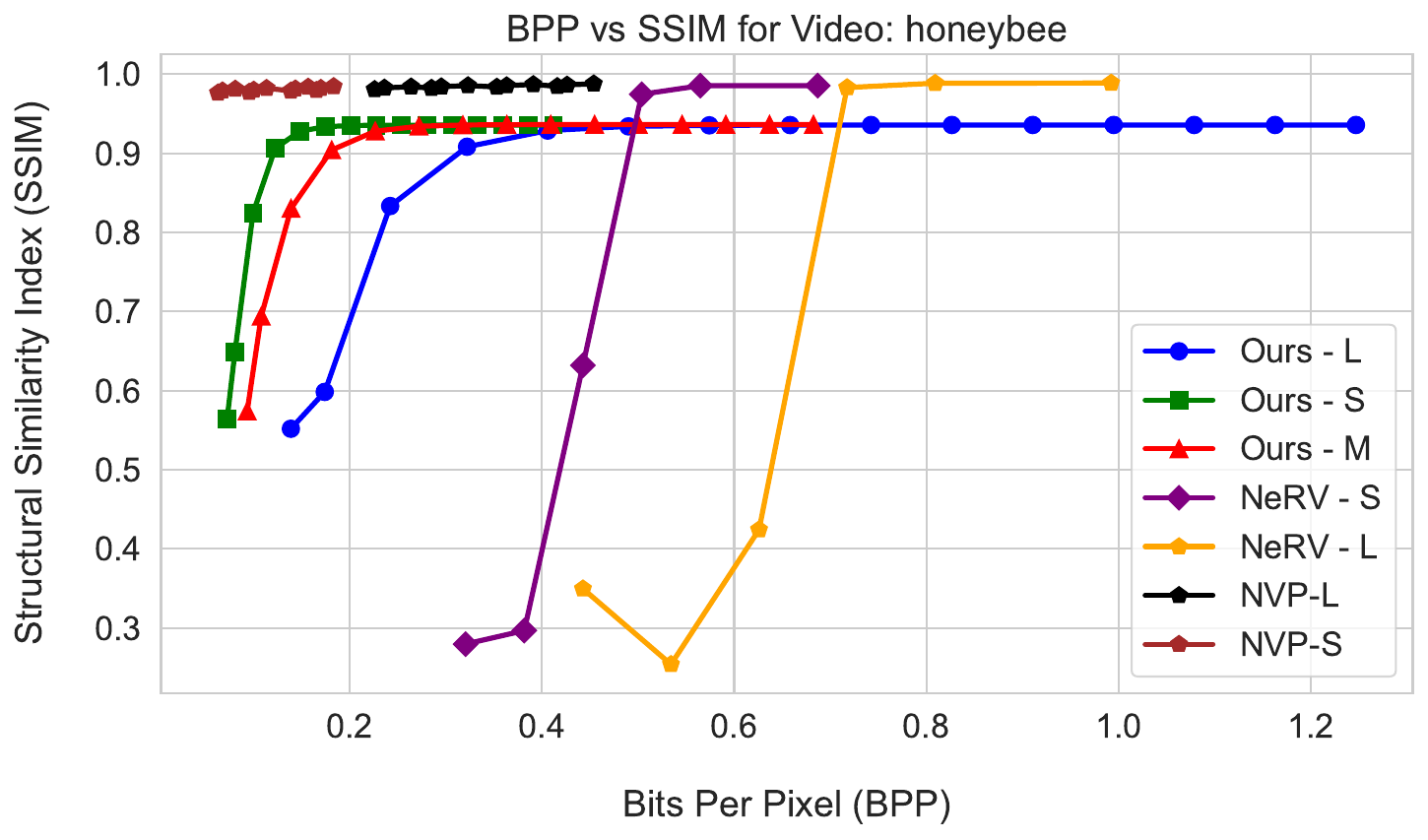}
    \end{subfigure}
    \caption{BPP vs. PSNR, SSIM for honeybee.}
    \label{fig:honeybee}
\end{figure}

\begin{figure}[ht]
    \centering
    \begin{subfigure}[b]{0.45\textwidth}
        \includegraphics[width=\textwidth]{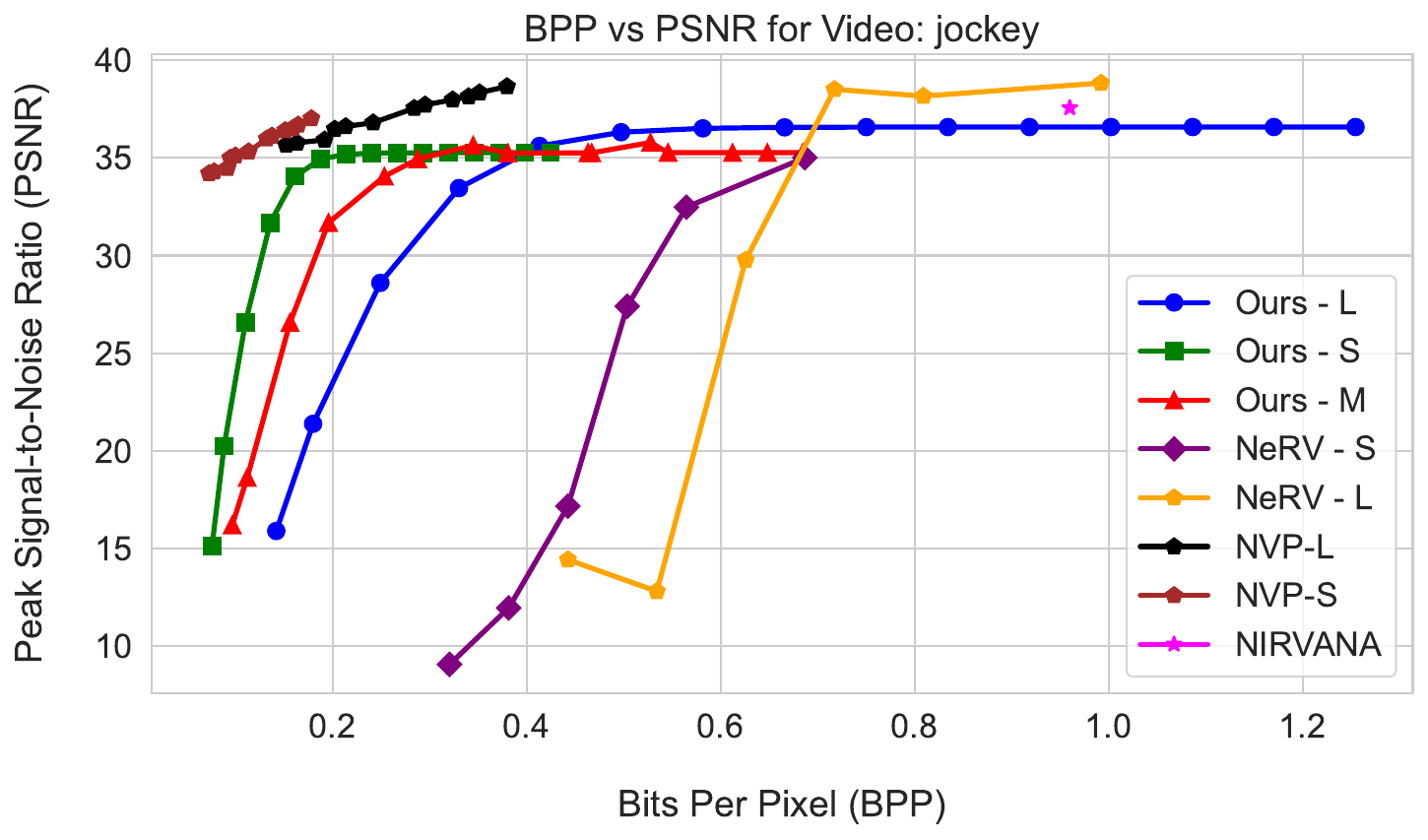}
    \end{subfigure}
    \begin{subfigure}[b]{0.45\textwidth}
        \includegraphics[width=\textwidth]{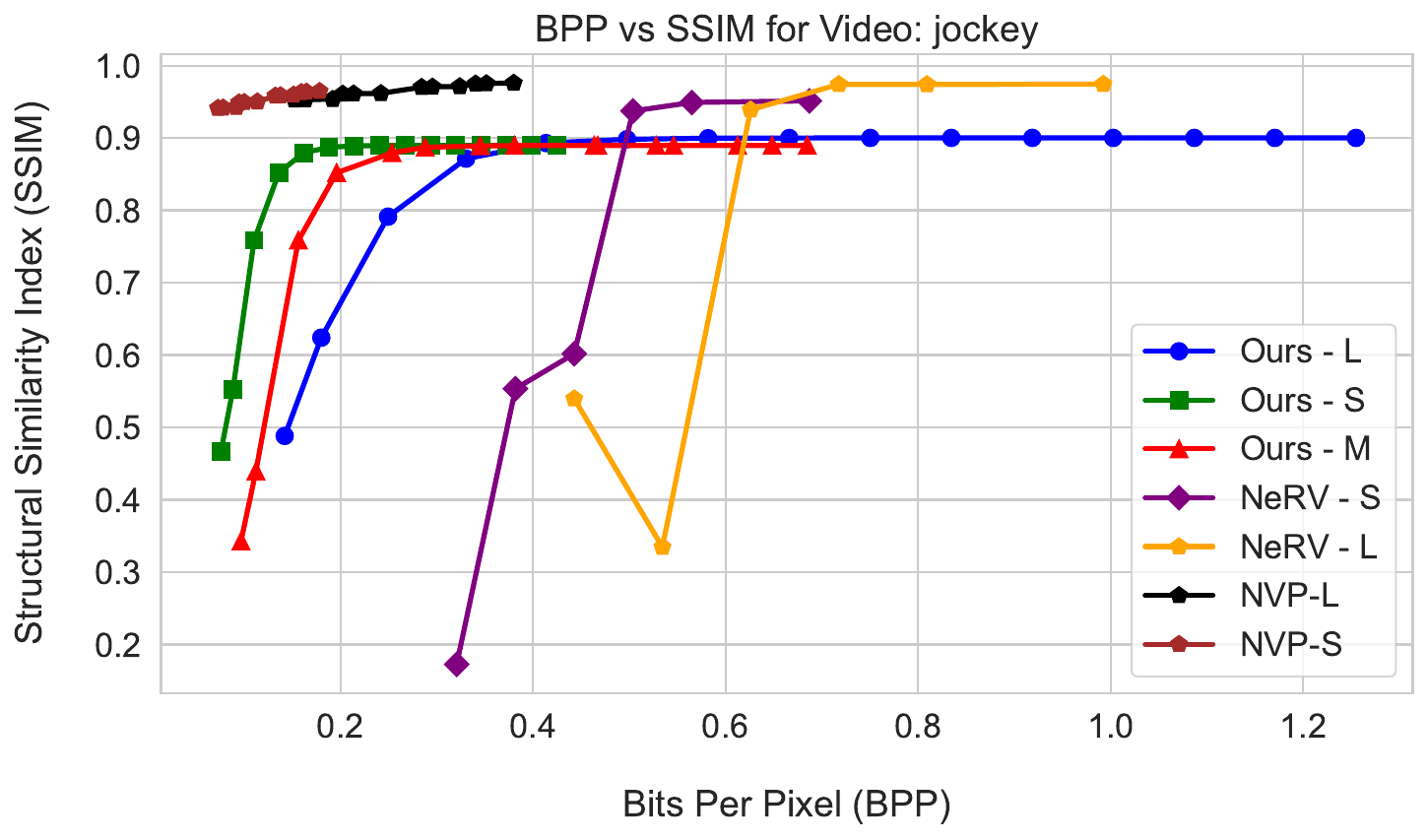}
    \end{subfigure}
    \caption{BPP vs. PSNR, SSIM for jockey.}
    \label{fig:jockey}
\end{figure}

\begin{figure}[ht]
    \centering
    \begin{subfigure}[b]{0.45\textwidth}
        \includegraphics[width=\textwidth]{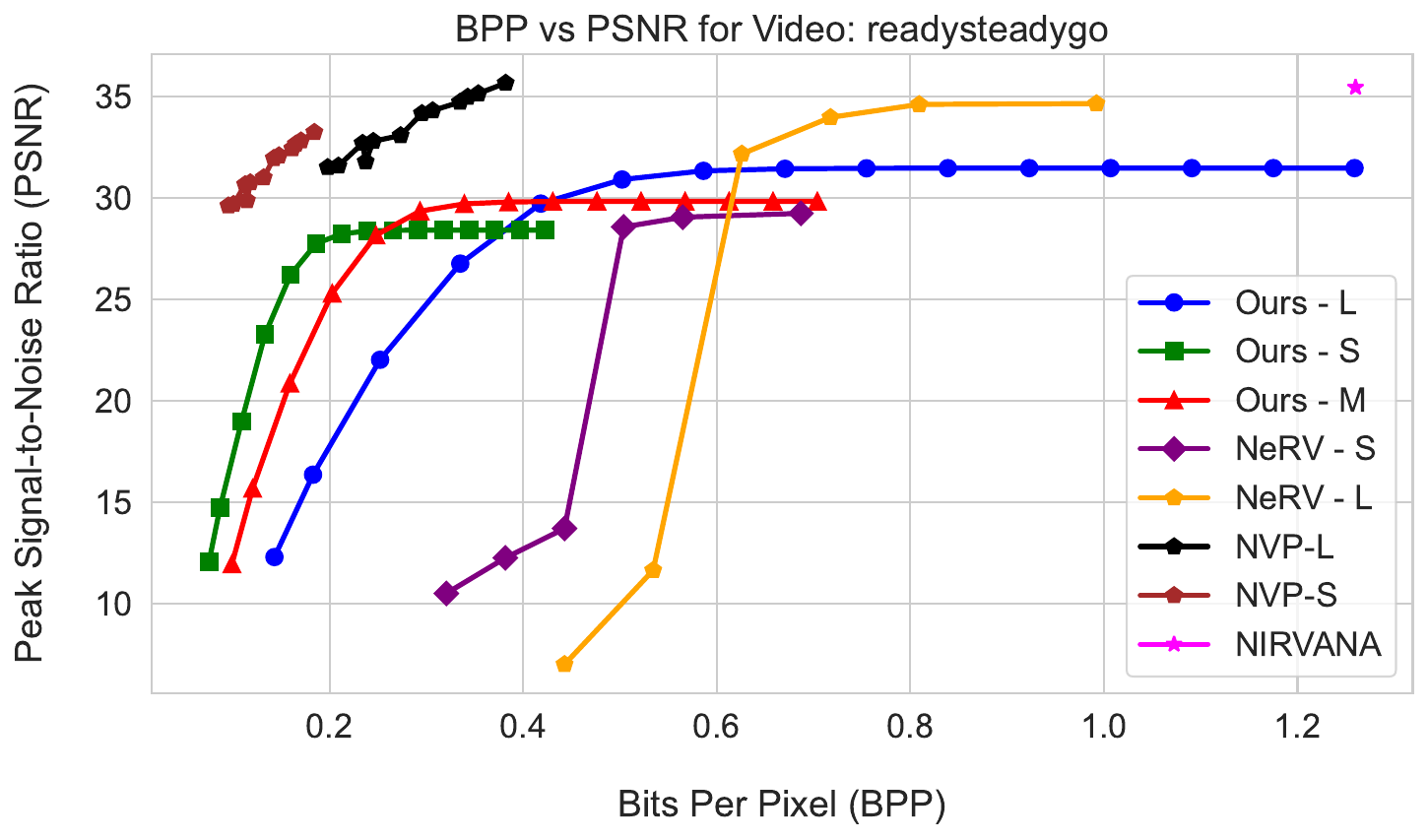}
    \end{subfigure}
    \begin{subfigure}[b]{0.45\textwidth}
        \includegraphics[width=\textwidth]{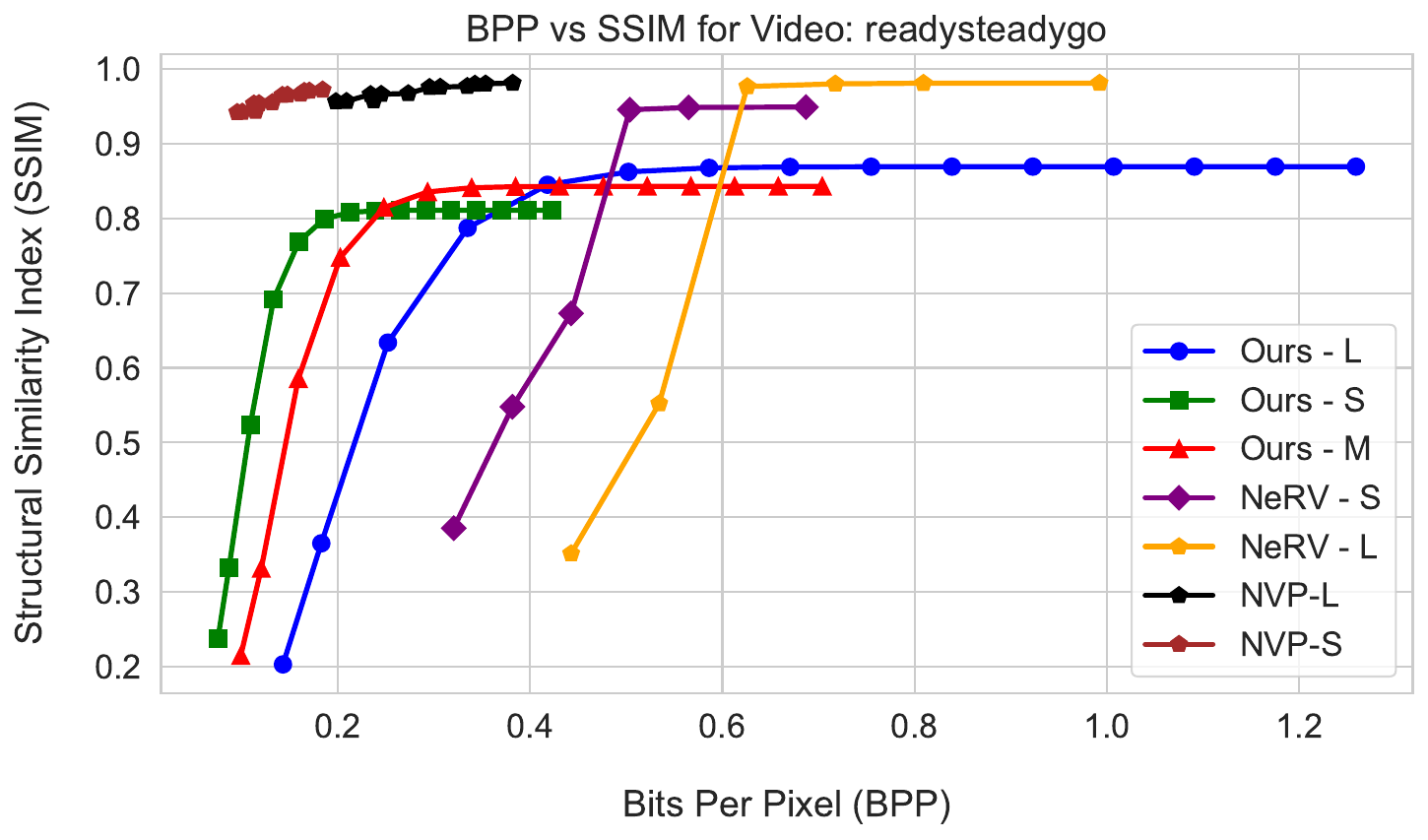}
    \end{subfigure}
    \caption{BPP vs. PSNR, SSIM for readysteadygo.}
    \label{fig:readysteadygo}
\end{figure}

\begin{figure}[ht]
    \centering
    \begin{subfigure}[b]{0.45\textwidth}
        \includegraphics[width=\textwidth]{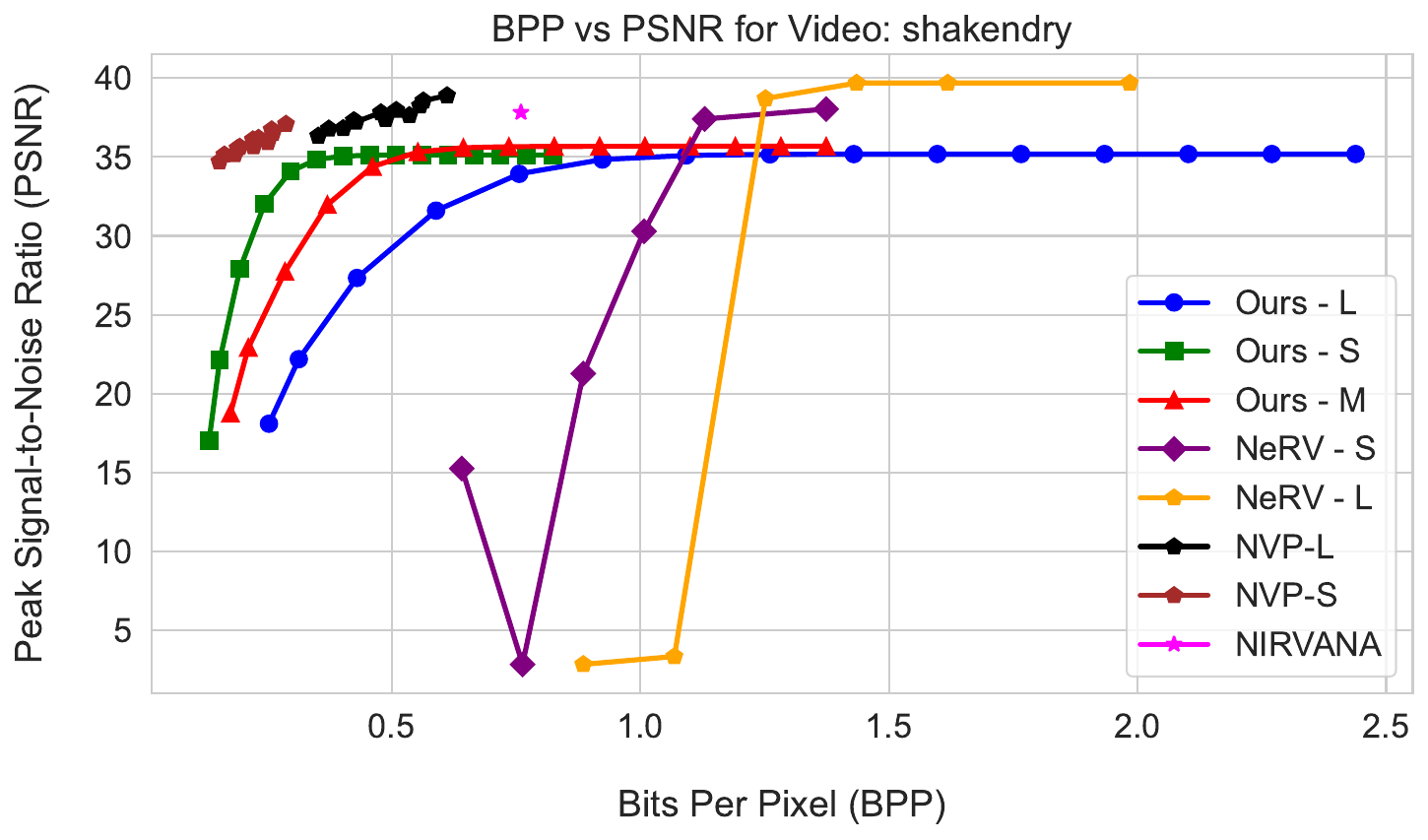}
    \end{subfigure}
    \begin{subfigure}[b]{0.45\textwidth}
        \includegraphics[width=\textwidth]{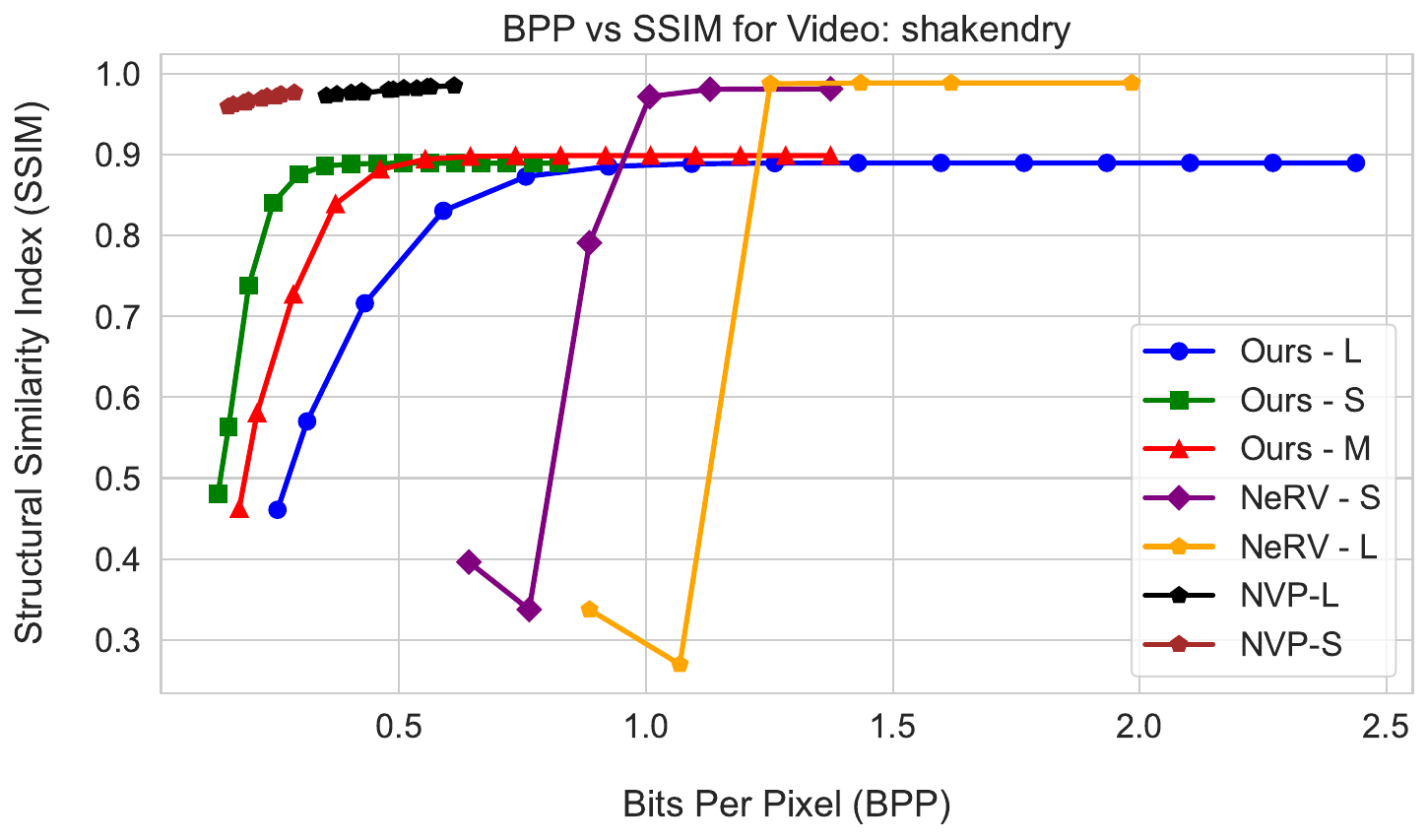}
    \end{subfigure}
    \caption{BPP vs. PSNR, SSIM for shakendry.}
    \label{fig:shakendry}
\end{figure}

\begin{figure}[ht]
    \centering
    \begin{subfigure}[b]{0.45\textwidth}
        \includegraphics[width=\textwidth]{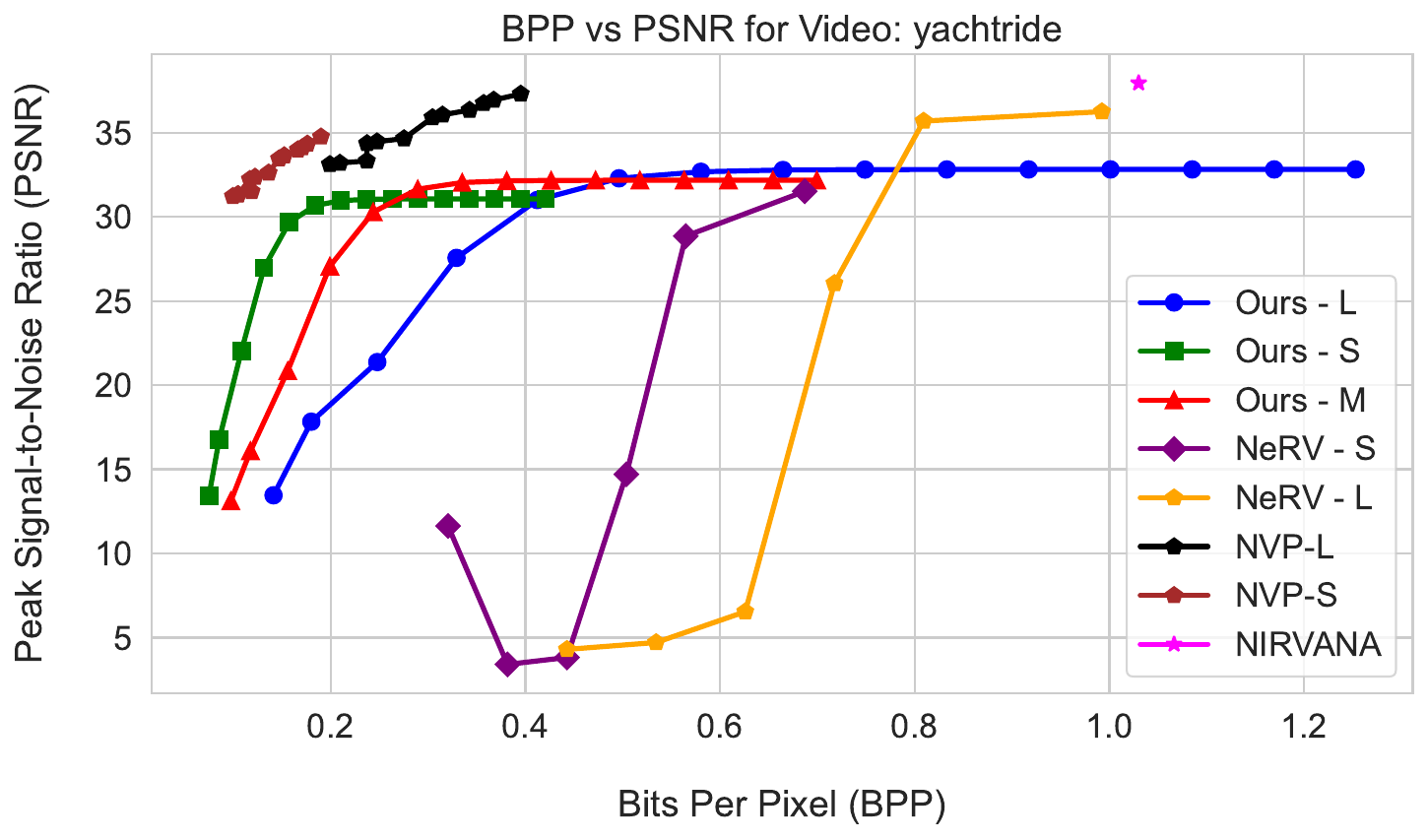}
    \end{subfigure}
    \begin{subfigure}[b]{0.45\textwidth}
        \includegraphics[width=\textwidth]{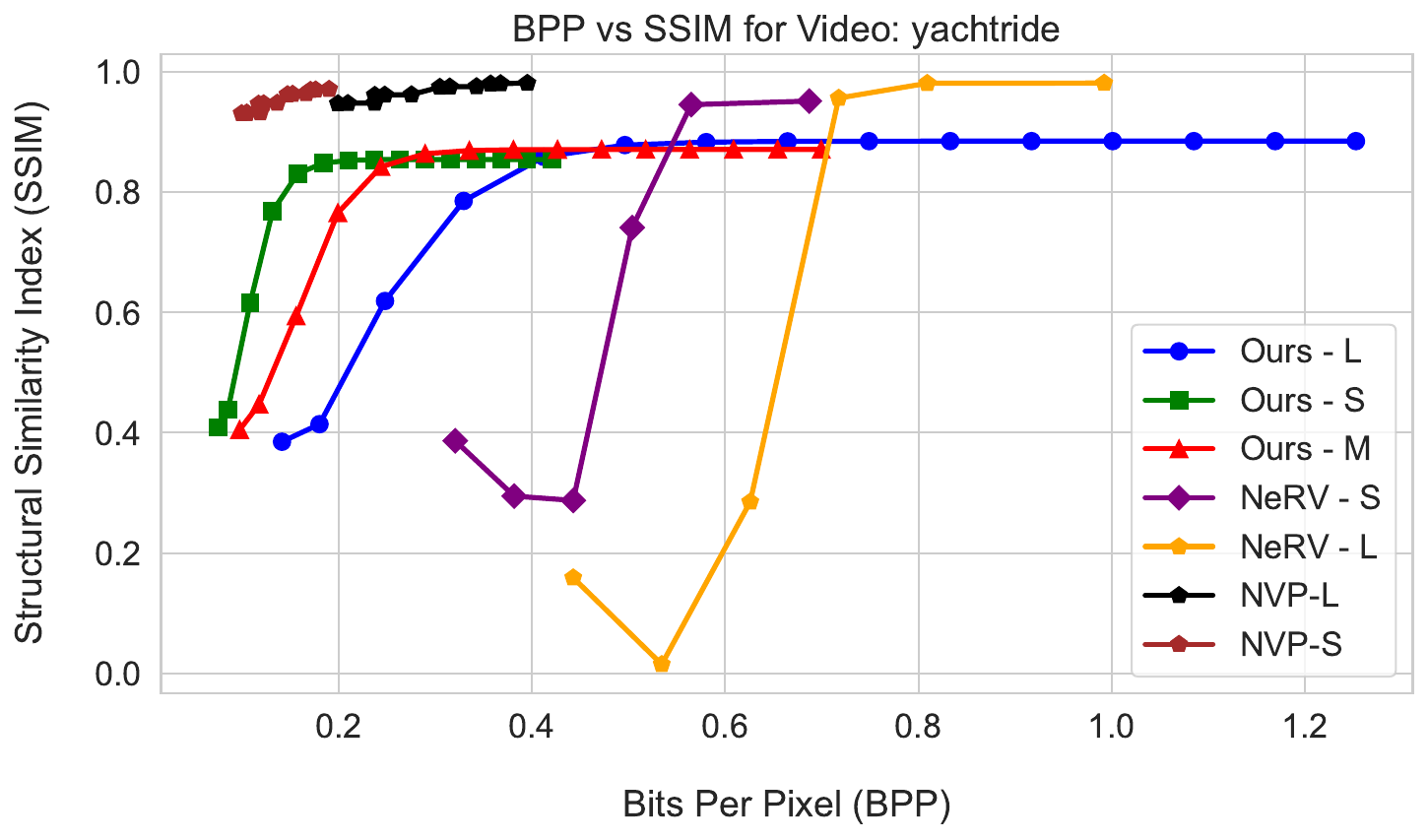}
    \end{subfigure}
    \caption{BPP vs. PSNR, SSIM for yachtride.}
    \label{fig:yachtride}
\end{figure}

We plot the results for each video from UVG dataset \cite{10.1145/3339825.3394937} in Figures~\ref{fig:beauty}, \ref{fig:bosphore}, \ref{fig:honeybee}, \ref{fig:jockey}, \ref{fig:readysteadygo}, \ref{fig:shakendry}, and \ref{fig:yachtride}.
We show three versions of our model based on the dimension of the  \textit{low-rank} modulating matrix. The \textit{Ours-s, Ours-m}, and \textit{Ours-l} correspond respectively to $size=50, 100, 200$ The \textit{Ours-m} model achieves reasonable performance when compared to other methods,  and at the same time can do the downstream tasks of interpolation and retrieval which none of the compared methods can.

\end{document}